\documentclass[nonblindrev]{workingpaperJJB} %

\usepackage[normalem]{ulem}
\usepackage{natbib}
\usepackage{amsmath,bbm}
\usepackage[capitalise]{cleveref}
 \bibpunct[, ]{(}{)}{,}{a}{}{,}%
 \def\bibfont{\small}%
 \usepackage{bibspacing}
 \TheoremsNumberedThrough     %
 \ECRepeatTheorems

 \EquationsNumberedThrough    %

\usepackage{amsfonts}
\usepackage{amsmath}  
\usepackage{amssymb}
\usepackage{changepage}
\usepackage{graphicx}
\usepackage{multirow}
\usepackage{enumerate}
\usepackage{enumitem}
\usepackage{booktabs}
\usepackage{color}
\usepackage{filecontents}
\usepackage{subcaption}
\usepackage[customcolors]{hf-tikz}
\usepackage{colortbl}
\usepackage{threeparttable}

\usepackage{algorithm}%
\usepackage{algpseudocode}%

\def\sym#1{\ifmmode^{#1}\else\(^{#1}\)\fi}

\newcommand{\added}[1]{{\leavevmode#1}}

\newcommand{\eps}{\varepsilon}
\newcommand{\bI}{\mathbbm{1}}
\newcommand{\bN}{\mathbb{N}}
\newcommand{\bR}{\mathbb{R}}
\newcommand{\bE}{\mathbb{E}}

\newcommand{\bS}{\mathbf{S}}
\newcommand{\bA}{\mathbf{A}}

\newcommand{\cI}{\mathcal{I}}

\newcommand{\cA}{\mathcal{A}}
\newcommand{\cS}{\mathcal{S}}

\newcommand{\tH}{\tilde{H}}

\newcommand{\nz}{\bar{z}}
\newcommand{\hz}{\hat{z}}
\newcommand{\yit}{\nz_{it}}

\newcommand{\htheta}{\hat{\theta}}

\newcommand{\Aspace}{\mathcal{A}_N}

\newcommand{\OPT}{\mathsf{OPT}}
\newcommand{\ALG}{\mathsf{ALG}}
\newcommand{\ONE}{\mathsf{ONE}}
\newcommand{\RAND}{\mathsf{RAND}}

\newcommand{\bandit}{\mathsf{Bandit}}
\newcommand{\QWI}{\mathsf{QWI}}
\newcommand{\baseline}{\mathsf{Baseline}}

\newcommand{\NULL}{\mathsf{NULL}}

\newcommand{\DPI}{\mathsf{DecompPI}}

\newcommand{\Bern}{\mathsf{Bern}}

\newcommand{\sumT}{\sum_{t=1}^T}
\newcommand{\sumN}{\sum_{i=1}^N}

\NewDocumentEnvironment{myproof}{o}
  {\IfNoValueTF{#1}{\paragraph{{Proof.} }} {\paragraph{{#1.} }} }
  {\hfill$\Halmos$}

\begin{document}

 \RUNAUTHOR{Baek et al.}

\RUNTITLE{Policy Optimization for Personalized Interventions}

\EquationsNumberedThrough    %

\TITLE{
Policy Optimization for Personalized Interventions in Behavioral Health}
\ARTICLEAUTHORS{
\AUTHOR{Jackie Baek}\AFF{Stern School of Business, New York University, \EMAIL{baek@stern.nyu.edu}}
\AUTHOR{Justin J. Boutilier}\AFF{Telfer School of Management, University of Ottawa, \EMAIL{boutilier@telfer.uottawa.ca}}
\AUTHOR{Vivek F. Farias}\AFF{Sloan School of Management, Massachusetts Institute of Technology, \EMAIL{vivekf@mit.edu}}
\AUTHOR{J\'{o}nas Oddur J\'{o}nasson}\AFF{Sloan School of Management, Massachusetts Institute of Technology, \EMAIL{joj@mit.edu}}
\AUTHOR{Erez Yoeli}\AFF{Sloan School of Management, Massachusetts Institute of Technology, \EMAIL{eyoeli@mit.edu}}
}
\HISTORY{\today}

\ABSTRACT{

Behavioral health interventions, delivered through digital platforms, have the potential to significantly improve health outcomes, through education, motivation, reminders, and outreach. We study the problem of optimizing personalized interventions for patients to maximize a long-term outcome, where interventions are costly and capacity-constrained. We assume we have access to a historical dataset collected from an initial pilot study. 
We present a new approach for this problem that we dub $\DPI$, which decomposes the state space for a system of patients to the individual level and then approximates one step of policy iteration.
Implementing $\DPI$ simply consists of a prediction task using the dataset, alleviating the need for online experimentation.
$\DPI$ is a generic model-free algorithm that can be used irrespective of the underlying patient behavior model.
We derive theoretical guarantees on a simple, special case of the model that is representative of our problem setting.
When the initial policy used to collect the data is randomized, we establish an approximation guarantee for $\DPI$ with respect to the \textit{improvement} beyond a null policy that does not allocate interventions. We show that this guarantee is robust to estimation errors. We then conduct a rigorous empirical case study using real-world data from a mobile health platform for improving treatment adherence for tuberculosis.
Using a validated simulation model, we demonstrate that $\DPI$ can provide the same efficacy as the status quo approach with approximately \textit{half} the capacity of interventions.
$\DPI$ is simple and easy to implement for an organization aiming to improve long-term behavior through targeted interventions, and this paper demonstrates its strong performance both theoretically and empirically, particularly in resource-limited settings.
}

\HISTORY{\today}

\maketitle

\section{Introduction}\label{s.intro}

For most health conditions, long-term outcomes are determined not only by clinical interventions but also by individual habits and behaviors \citep{bosworth2011medication}. A range of behavioral health interventions, often delivered through digital platforms, have been shown to improve outcomes for various health conditions. These interventions aspire to affect users' habits through motivation, education, nudging, or boosting \citep{Ruggeri20Behavioral}. At the same time, such interventions are associated with a variety of direct and indirect costs. This paper is concerned with optimizing the impact of costly interventions through prioritization. 

As a concrete example of a digital health intervention, consider our partner organization, \textit{Keheala}. Keheala operates a digital service promoting medication adherence among patients prescribed with tuberculosis (TB) treatment in Kenya, and has already been shown to have a non-trivial impact on adherence and health outcomes \citep{Yoeli19Digital,Boutilier22Improving}. 
Increasing adherence to the prescribed course of treatment for TB is important since lack thereof can result in a relapse of the disease, and worse yet, the emergence of multi-drug resistant strains of the TB bacteria. A key feature of Keheala's adherence program is that patients are required to self-verify treatment adherence daily via a mobile phone interface. In addition, Keheala's digital platform comprises of a suite of interventions to further support adherence. Some are on-demand (e.g., leaderboards and information about TB) and some are automated (e.g., daily reminders to adhere to treatment). Most importantly, some are manual (e.g., outreach messages and phone calls). Keheala employs \textit{support sponsors} to operate the platform and reach out to patients who have not self-verified adherence for a predetermined amount of time. The interactions between the platform and the patients result in a feedback loop that presents an opportunity for Keheala to discern the impact of each type of intervention as well as to identify patients most in need of outreach.

The Keheala case highlights three salient features that are common among digital services in the context of behavioral health services. 
\begin{enumerate}
	\item Interventions are costly and must be rationed. In general, even automated mobile reminders entail an indirect cost, in that frequent reminders are likely to sensitize patients to the automated messages. In the case of Keheala, calls from support sponsors entail a direct cost. Due to these costs, demand for support sponsor outreach exceeds supply. The support sponsors had a range of responsibilities but on an average day in our data (see more details in \cref{s.casestudy}) their effective capacity to make phone calls to patients was exceeded by the number of patients eligible for such phone calls by a factor of 8.
	\item The metric of interest is a binary measure of compliance among users. For many behavioral health applications, a service provider (e.g., digital platform) collects data from users, aimed at identifying whether they are in a state of compliance with a desired behavior (e.g., medication adherence, exercise routine, correct diet) or not. In the case of Keheala, this binary outcome measure is the daily self-verification of treatment adherence. 
	\item \added{Initially, limited data is collected using an ad hoc baseline policy. Most digital behavioral health services are initially launched (e.g., as part of a pilot study) with heuristic rules of thumb guiding the timing of outreach to each user \citep{Mills20Personalized}. These heuristics could be in the form of binary eligibility criteria (as opposed to an ordered ranking) which do not allow for prioritization among eligible users during periods when their numbers exceed the available capacity. As a result, a subset of eligible users end up receiving outreach whereas others do not. In Keheala's case, an RCT was conducted in which the protocol simply stated that patients who had not self-verified their adherence for 48 hours should be prioritized for support sponsor outreach.}
\end{enumerate}

This state of affairs prompts the key question we seek to answer: {\em Can one use limited pilot-study data, collected using some ad hoc baseline policy, to design a practical prioritization policy that maximizes the impact of costly interventions?} 

\subsection{What can Reinforcement Learning Offer?}
Before diving into the development of a new approach to the problem, let us explore an out-of-the-box approach for this problem. 
Natural candidates for personalizing outreach interventions using data are reinforcement learning (RL) algorithms. 
As a prototype of the sorts of RL approaches that one might apply to intervention optimization, consider formulating the task at hand as a so-called \textit{contextual bandit}. Specifically, the context of a given patient is their adherence and intervention history so far, and the reward corresponds to whether or not they self-verify treatment adherence in the following day. This reward is assumed to be some unknown, noisy function of context and the chosen action (i.e., whether or not the patient received an intervention). 
Of course, a myopically optimal action may be sub-optimal, but let us ignore this limitation for now. 

\begin{figure}[h]
	\begin{center}
		\vspace{-2mm}
		\includegraphics[width=0.75\linewidth]{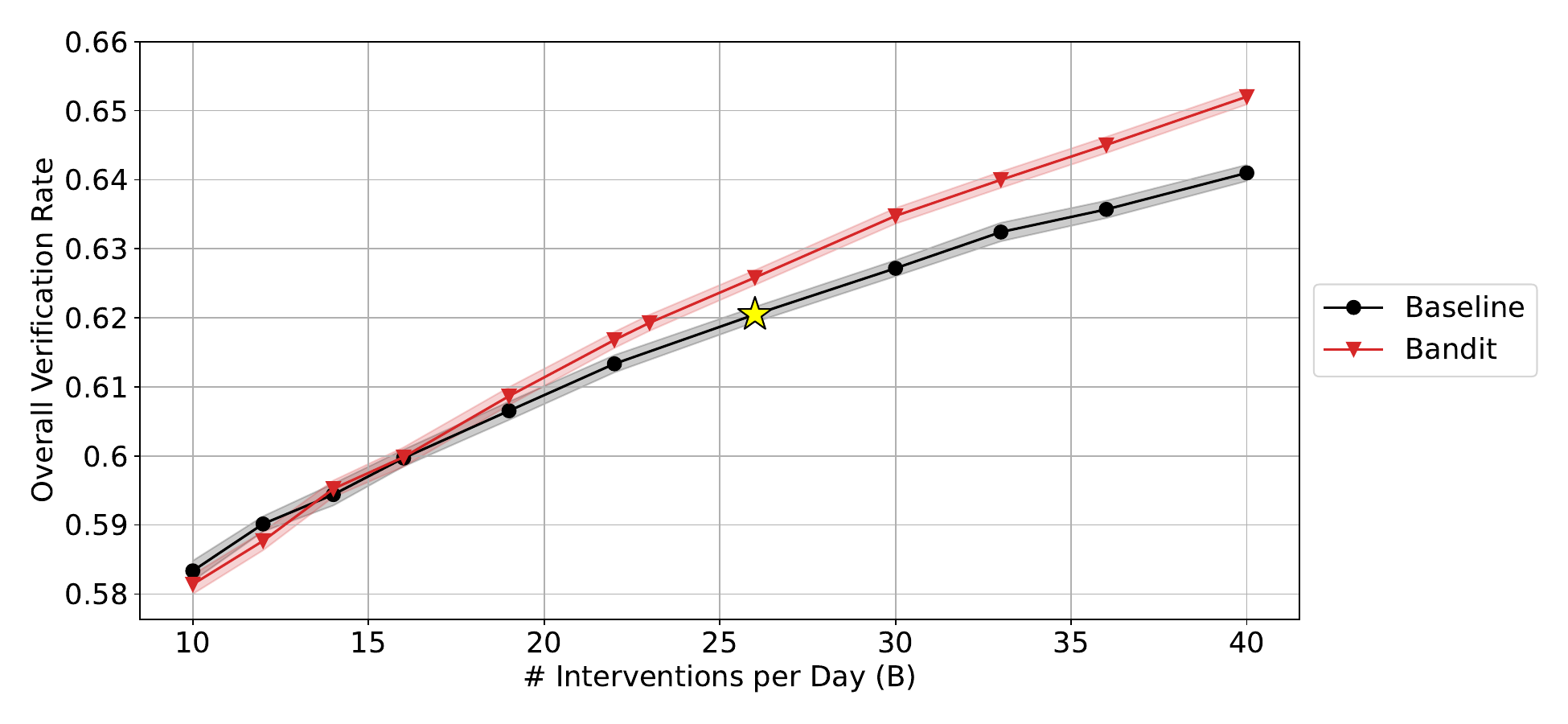}
		\vspace{-2mm}
		\caption{Average overall verification rate over 50 runs for each policy and budget. The shaded region indicates a 95\% confidence interval. The star represents the operating point for Keheala.}
		\label{fig:introfig}
		\vspace{-6mm}
	\end{center}
\end{figure}

Figure \ref{fig:introfig} compares the performance of a common contextual bandit algorithm, Thompson Sampling, with that of the existing heuristic employed by Keheala using a simulation model fit to data from Keheala's initial pilot.\footnote{The model itself was trained by cross-validation with double ML and further validated on a hold out set; see \cref{s.casestudy} for details on this as well as the bandit approach itself.} In parsing this figure, we note that the x-axis corresponds to the number of calls (and thus, also the amount of data that can be collected about the incremental impact of an intervention) in a single day. 
The operating point for the Keheala pilot corresponds to a budget of 26 interventions per day, as indicated by the yellow star.

While we acknowledge that a contextual bandit is only one simple approach to our problem (we will include more sophisticated benchmarks in our numerical analysis in \cref{s.casestudy}) it is illustrative to highlight some of its shortcomings, relative to the criteria we had listed above. Setting aside the obvious issue that a contextual bandit relies on online exploration (whereas we would like to learn from offline data), we make three additional observations. 
First, in general we observe modest improvements. The bandit captures a similar verification rate as the incumbent Keheala policy with 3 fewer interventions per day (relative to the 26 used by Keheala). While meaningful, this is a somewhat marginal improvement.
Second, performance can actually deteriorate, relative to the baseline policy, particularly in the most resource-limited cases. As we further discuss in \cref{ss.results}, this is a data scarcity issue: with a small budget, the bandit simply does not collect enough exploration data to learn an improved policy. This is despite the fact that the Thompson sampling algorithm employed had the benefit of a prior learned on approximately half of the data. 
Third, the contextual bandit is (by design) a myopic approach. In essence, the bandit learns to provide outreach in such a way as to maximize the next day's self-verification rates, which may be highly suboptimal for the long term. While, we could certainly turn to a reinforcement learning algorithm that attempts to learn an optimal policy by measuring the long run impact of an action, the data requirements of such a policy would increase substantively so that the performance degradation we see for the bandit at low budgets is likely to persist at (much) higher budgets.

In summary, this leaves us in a peculiar spot---whereas more sophisticated RL algorithms could, in principle, learn an optimal policy, we typically do not have the data budget, or the willingness to risk policy degradation due to exploration, to use such approaches. However, simpler RL algorithms (such as the contextual bandit) provide a relatively marginal improvement.

\subsection{This Paper}
In this paper, we propose and evaluate a new algorithm for intervention optimization problems of the type described above. 

Our first contribution is the algorithm itself (\cref{s.model}). Our starting point is a formal model of the rich practical setting of interest where we model patient behaviors using a generic  Markov Decision Process (MDP)---we refer to this as the `full model'. We propose a new approach---inspired by existing techniques---for intervention optimization problems we dub \textit{Decomposed Policy Iteration $(\DPI)$}. Loosely, $\DPI$ can be thought of as approximating a single step of policy iteration in the high dimensional MDP corresponding to our problem. 
Taking as input a dataset of interventions and outcomes under an incumbent policy (also called baseline policy), 
we estimate the state-action values ($q$-values) of this policy using a prediction algorithm.
Then, at each time step, interventions are assigned to the patients who have the largest increase in their $q$-value if they receive an intervention, compared to if they do not.
Notably, the $q$-values are decomposed at the patient level, which avoids any dependence on an exponentially sized state space.
$\DPI$ is a promising practical solution to intervention optimization problems as it is model-free (i.e., does not require access to a model of the environment), and
it also does not rely on online experimentation or updating, as is the case for many RL algorithms.
\added{We establish that $\DPI$ will improve upon the baseline policy when the baseline policy lies in a class of random policies.}

Our second contribution is to provide surprisingly strong performance guarantees for $\DPI$ when applied to a stylized version of the full problem (\cref{s.theory}). 
The `full model' from \cref{s.model} can model \emph{arbitrary} MDPs. 
However, patient dynamics are not completely arbitrary; in Keheala, we observe that patients exhibit \emph{habitual} behavior.
We therefore study the performance of our algorithm when patient behavior is driven by a standard simple model\footnote{We emphasize that $\DPI$ is a generic model-free algorithm that can be used irrespective of the underlying patient behavior model. The behavioral model is used only for the theoretical analysis.} akin to models considered both in the behavioral health \citep{mate2020collapsing,mate2022field,biswas2021learn} and marketing literature \citep{schmittlein1987counting}. For situations in which user behavior is driven by such models, we establish an approximation ratio of the {\em improvement} between an optimal policy and a null policy that does not allocate interventions. 
Notably, this result is much stronger than a typical approximation guarantee, which compares the \textit{absolute} performance of a policy to the optimal policy. 
Specifically, when the baseline policy used to collect the data is randomized, the approximation ratio of the improvement approaches $1/2$ as the intervention capacity decreases.
Since the $q$-values used by $\DPI$ need to be estimated, we prove that our guarantee is robust in the sense that the performance guarantee scales gracefully with respect to errors in these quantities.

Our third contribution is to return to the full model and provide numerical results based on field data (\cref{s.casestudy}). Using the validated simulation setup described in Figure~\ref{fig:introfig}, we find that $\DPI$ would yield the same verification rate as the incumbent Keheala policy {\em with less than half the capacity}. Furthermore, $\DPI$ outperforms the contextual bandit algorithm (described above) for budgets ranging from 50\% to 150\% of the original capacity. Importantly, the performance gains are highest in low capacity scenarios---the most likely paradigm for a future large-scale roll-out of the system in resource-limited settings. This is encouraging for digital behavioral health applications, more generally---particularly given how well the policy performs despite only having access to limited data collected during pilot implementation.

\section{Literature Review}\label{s.litreview}

Our work relates to the expansive streams of literature on reinforcement learning and (approximate) dynamic programming, as well as the applied operations research literature focusing on improving healthcare delivery in resource-limited settings. Methodologically speaking, most existing solution approaches to problems similar to ours can be classified as either (a) developing policies for a \textit{known} underlying model of behavior or (b) \textit{learning} a policy using data. We summarize these two streams of work in Sections \ref{ss.knownmod} and \ref{ss.unknownmod}, respectively, before discussing prior work on TB treatment as an application area in \cref{ss.litreviewhealth}.

\subsection{Known Model}\label{ss.knownmod}
Our model (described in detail in \cref{s.gmodel}) assumes that every patient behaves according to a Markov decision process (MDP). Even if all the MDP parameters of these models were known exactly, the size of the system state space would be \textit{exponential} in the number of patients: the size of the state space is $|\cS|^N$, where $\cS$ is the state space for one patient and $N$ is the number of patients. As a result, a direct application of dynamic programming techniques such as backwards induction or policy/value iteration would take exponential time and hence is practically infeasible.

As exact methods are infeasible, one can resort to \textit{approximate} dynamic programming (ADP) techniques developed for weakly-coupled MDPs \citep[e.g.,][]{meuleau1998solving,adelman2008relaxations,d2023optimal}. Specifically, the model we study is a \textit{restless bandit}: each patient corresponds to an arm, and there is a budget on the number of arms that can be `pulled' (assigned an intervention) at each time step. The state of each arm evolves as a Markov chain, where its transition probabilities depend on the action taken. It is known that finding the optimal policy to a restless bandit is PSPACE-hard \citep{papadimitriou1994complexity}. There is a large literature on developing algorithms for this problem \citep[e.g., ][]{whittle1988restless,glazebrook2002index,ansell2003whittle,glazebrook2006some,liu2010indexability,guha2010approximation}.
A commonly used policy is the Whittle's index \citep{whittle1988restless}, which is known to be asymptotically optimal under certain strong conditions \citep{weber1990index} and has been shown to have good empirical performance 
\citep{ansell2003whittle,glazebrook2002index,glazebrook2006some}.
However, the Whittle's index have mostly been applied to problems with a single-dimensional parameter, as the computation of the Whittle's index for a single arm is cubic in the number of states \citep{nino2020fast}.
Therefore, even when the parameters are known exactly, the restless bandit is far from a solved problem, especially when dealing with a large or infinitely-sized state space.

\subsection{Unknown Model}\label{ss.unknownmod}
Deriving an optimal policy for an MDP with unknown parameters corresponds to reinforcement learning (RL), a rapidly expanding area of research \citep{sutton2018reinforcement}.
However, a naive application of the RL framework onto our problem results in an exponentially large state space --- this correspondingly results in an exponential blowup in the data requirements to deploy generic RL methods such as Q-learning \citep{jin2018q}.
\added{
Further, we consider the more difficult \emph{offline} RL problem, where we assume access to a dataset collected from another policy and no further access to the environment 
--- see \cite{levine2020offline} for a survey of offline RL (note that this literature studies the generic RL problem, not the restless bandit setting).
In this literature, \cite{brandfonbrener2021offline} shows that performing one step of policy iteration exhibits strong empirical performance, compared to other iterative methods that rely on off-policy evaluation.
Our algorithm also performs one step of policy iteration, but we introduce an additional modification where we \textit{decompose} the value function by patient, to address the issue of the exponential state space.
}

\paragraph{Greedy and multi-armed bandits.}
One approach, as described in the introduction, is to greedily maximize the immediate reward at each time step.
For example, in the treatment adherence setting, one can reach out to patients with the highest increase in their probability of adhering in the next day from the intervention.
A multi-armed bandit is one natural framework that can be used to learn such a policy.
This is the approach used in HeartSteps \citep{lei2017actor,liao2020personalized}, a program to promote physical activity using real-time data collected from wristband sensors.
These papers use a contextual bandit model, where the context represents a user at a particular time step, and they develop a bandit algorithm whose goal is to increase the immediate activity level of the user.
Given the vast literature on contextual bandits, there are a wide variety of algorithms that one can easily plug in.

A fundamental downside of this greedy approach is that it does not capture any potential long-term effects of an action---that is, an action may not only impact a patient's immediate behavior, but their behavior for all future time steps.
One way to address this is to specifically model the type of long-term effect it can have.
For example, \cite{liao2020personalized} introduce a `dosage' variable that models the phenomenon that the treatment effect of an action is often smaller when an action was recently given to that patient in the past.
Similarly, \cite{mintz2020nonstationary} incorporate habituation and recovery dynamics into the bandit framework.
However, these approaches capture only particular types of long-term effects that are explicitly modeled, and there could be other, complex factors that affect the downstream behavior of patients.

Our work does not take this greedy approach, and we aim to learn a policy that maximizes long-term rewards, without specifically modeling the type of long-term effect that an action can have.
We benchmark against a contextual bandit policy, and we observe that incorporating the long-term effects is critical in the behavioral health setting that we study. 

\paragraph{Learning for restless bandits.}
A non-greedy approach to this problem corresponds to the restless bandit model in the unknown parameter regime.
There is a nascent literature develops learning algorithms in this setting. One approach is to adapt algorithms from the multi-armed bandit literature such as UCB \citep{wang2020restless} or Thompson Sampling \citep{jung2019regret}. 
Recent approaches similarly adapt reinforcement learning methods such as Q-learning \citep[e.g., ][]{fu2019towards,avrachenkov2022whittle}. Importantly, these methods are \textit{online} learning algorithms that require continuous exploration and assume that the state space is known and finite. This literature focuses on providing theoretical guarantees of the proposed algorithms, and these algorithms eventually converge to the optimal policy.

Our work differs from the aforementioned literature in \added{four main} ways.
First, we take an \textit{offline} approach, where we leverage existing data to derive a new policy.
This removes the need for exploration, as well as the dependence on a long horizon to obtain an improvement over a baseline policy.
Second, our work does not focus on deriving an \textit{optimal} policy; we derive a practical policy that can be implemented with limited data, and our theoretical results are approximation guarantees to the optimal policy.

\added{Third, we take a \textit{model-free} approach, rather than the commonly used \textit{model-based} approach in restless bandits, in which a simple model of patient behavior is postulated using a small state space.}
For example, \cite{mate2022field} and \cite{biswas2021learn} posit a simple MDP with two and three states respectively for each user, where a state represents the engagement level of the user. 
\cite{aswani2019behavioral} take a slightly different model-based approach in studying weight loss interventions, where they model user behavior via utility functions.
Then, the policy is developed based on these posited models, and therefore these approaches rely heavily on the correctness of the model.
In contrast, our model-free approach is to incorporate as much information as available into the patient's state (which leads to a relatively large state space), and our policy then operates under the assumption that patients in similar states will behave similarly.
\added{Such an approach relies on being able to easily handle a very large, or infinitely-sized state space, which is not the case for for classic restless bandit policies like the Whittle's index.
Our approach is affected by the state space only through a prediction task, which can be performed with any machine learning algorithm (which are usually designed to be able to handle a large state space).
}
In \cref{s.theory}, we use a simple two-state MDP for the purposes of proving a theoretical guarantee for our policy, but the policy is defined irrespective of the underlying patient behavior model.

\added{
Lastly, we note that our technical analysis does not build off of proof ideas from the RL literature --- rather, it borrow ideas from the literature on the online allocation of reusable resources (which, on the surface, seems unrelated to our problem setting).
The key connection relies on interpreting one \emph{patient} as one \emph{reusable resource}. Since we model one patient as being in one of two states (0 or 1), 
we can interpret a patient being in state 0 and state 1 as a reusable resource being \emph{available} and \emph{unavailable} to sell, respectively.
As patients can transition from state 1 to 0, this is analogous to a sold resource returning to the available state.
Our analysis for approximation guarantees builds off of a proof technique used in \cite{gong2021online}.
}

\subsection{Healthcare systems} \label{ss.litreviewhealth}
Finally, from an application perspective, our work contributes to a growing literature focusing on improving healthcare delivery systems in resource-limited settings. Most related to the paper at hand are recent papers focusing on improving TB outcomes in resource-limited settings. Much of this work has been on the policy level, with \cite{Suen14Disease} evaluating strategic alternatives for disease control of multi-drug resistant TB (MDR TB) in India and finding that with MDR TB transitioning from treatment-generated to transmission-generated, rapid diagnosis of MDR TB becomes increasingly important. Similarly, \cite{Suen18Optimal} optimize the timing of sequential tests for TB drug resistance, a necessary step for transitioning patients to second-line treatment. 

Two papers focus on medication adherence. \cite{Suen22Design} tackle the problem of designing patient-level incentives to motivate medication adherence, in situations where adherence is observable but patients have unobserved and heterogeneous preferences for adherence. They first take a modeling approach to design an optimal incentive scheme and then demonstrate that deploying it would be cost effective in the context of TB control in India. Similar to us, \cite{Boutilier22Improving} focus on a behavioral intervention, demonstrating that data describing patient behavior (e.g., patterns of self-verification of treatment adherence, like in the case of Keheala) can be leveraged to predict daily behavioral outcomes as well as health outcomes. They use such predictions to assign patients to risk groups and demonstrate empirically that outreach can be effective, even for patients who are classified as at risk. However, they stop short of prescribing an actionable policy for assigning patient outreach, which is the topic of this paper.

\section{Full Model and Policy}\label{s.model}
We first formally describe the intended problem setting in its full complexity in \cref{s.gmodel}. 
We then describe our algorithm in \cref{s.decomposedPI} and discuss the intended usage and design choices of the algorithm in \cref{s.intended_usage}.
\added{
In \cref{ss.general_guarantees}, we show, using a counterexample, that our algorithm does not always guarantee an improvement over the baseline policy, but we then demonstrate that an improvement is guaranteed when the baseline policy belongs to a specific class of random policies. All proofs for this section and the next are relegated to appendices.}

\subsection{Model} \label{s.gmodel}
There are $N$ patients and $T$ time steps.
Each patient $i \in [N]$ is associated with a Markov decision process (MDP) represented by $\mathcal{M}_i = (\cS, \cA, \mathcal{P}_i, R_i)$.
$\cS$ is the state space, and let $S_{it} \in \cS $ denote the state of patient $i$ at time $t$.
The action space is $\cA = \{0, 1\}$, where we will refer to action $A_{it} = 0$ as the \textit{null} action, and action $A_{it}=1$ as the \textit{intervention}.
$\mathcal{P}_i(S, S', A) = \Pr(S_{i,t+1} = S' \;|\; S_{it} = S, A_{it} = A)$ is the probability of transition from state $S$ to $S'$ when action $A$ is taken, and $R_i(S, S', A)$ is the immediate reward from this transition.
If $A_{it} = 1$, we say that patient $i$ is \textit{chosen} at time $t$.

Next, we define a \textit{system MDP} by combining the $N$ patient MDPs and adding a budget $B \geq 0$ on the number of interventions that can be given at each time step.
That is, $\Aspace = \{(A_1, \dots, A_N) \subseteq \{0, 1\}^N: \sumN A_i \leq B\}$ is the action space for the system MDP.
The state space is $\cS^N$, 
and a policy $\pi = \{\pi_t \;|\; t \in [T]\}$ maps each state to a distribution over the action space $\Aspace$.
At each time $t = 1, \dots, T$, the following sequence of events occur:
\begin{enumerate}
	\item The state $S_{it} \in  \cS$ is observed for each patient $i \in [N]$.
	\item An action $(A_{1t}, \dots, A_{Nt})$ is drawn from $\pi_t(S_{1t}, \dots, S_{Nt})$.
	\item For each patient $i$, their next state $S_{i,t+1}$ is realized independently according to their transition probabilities $\mathcal{P}_i(S_{it}, \cdot, A_{it})$. We gain the reward $\sumN R_i(S_{it}, S_{i,t+1}, A_{it})$.
\end{enumerate}
Let $\mathbf{S}_1 = (S_{11}, \dots, S_{N1})$ be the starting state profile.
An instance of this problem is represented by $\cI = (N, T, (\mathcal{M}_i)_{i \in [N]}, B, \mathbf{S}_1)$.
We use the name of a policy to denote the expected total reward of that policy.

\subsection{Decomposed Policy Iteration} \label{s.decomposedPI}

We introduce a natural policy closely related to the ubiquitous \textit{policy iteration} (PI) algorithm \citep{howard1960dynamic}.
As the name suggests, policy iteration is an iterative algorithm that maintains a policy and updates it at each iteration to improve its performance.
Our policy, $\DPI$, can be thought of as performing \textit{one step} of policy iteration.
In addition, $\DPI$ differs from vanilla PI in that the algorithm decomposes to the patient level, which allows us to remove dependence on the exponentially sized state and action space.

\subsubsection{Intervention values.}
For any policy $\pi$, let $S^\pi_{it}$ and $A_{it}^\pi$ be the induced random variables corresponding to the state and action, respectively, of patient $i$ at time $t$ under policy $\pi$.
Then, we define 
\begin{align} \label{eq:q}
q_{it}^\pi(S, A) = \bE_{\pi}\left[\sum_{t'=t}^T R(S_{it'}^\pi, S_{i,t'+1}^\pi, A_{it'}^\pi) \;\bigg|\; S_{it}^\pi = S, A_{it}^\pi = A \right],
\end{align}
which represents the expected future reward from patient $i$ when running policy $\pi$, conditioned that they were in state $S$ and were given action $A$ at time $t$.
In the case where $\Pr(S_{it}^\pi = S, A_{it}^\pi = A) = 0$, we define $q_{it}^\pi(S, A) = 0$.
Next, we define an \textit{intervention value}, $z^{\pi}_{it}(S)$, to be the difference in the $q^{\pi}_{it}$ values under $A=1$ and $A=0$:
\begin{align*}
z^\pi_{it}(S) = q_{it}^{\pi}(S, 1) - q_{it}^{\pi}(S, 0).
\end{align*}
$z^\pi_{it}(S)$ is the difference in expected total future reward from patient $i$ under policy $\pi$ when the patient is given the intervention, compared to when they are not.

\subsubsection{Policy.}
Given a \textit{baseline policy} $\pi$, $\DPI(\pi)$ is the policy that gives the intervention to the patients with the highest positive intervention values, up to the budget constraint.
Formally, 
\begin{align*}
\DPI(\pi)_t(S_{1t}, \dots, S_{Nt}) \in \argmax_{A \in \cA_N} \sum_{i \in [N]} A_{i} z^\pi_{it}(S_{it}).
\end{align*}

\subsection{Intended Usage and Design Choices} \label{s.intended_usage}
Recall that our motivating research question is to leverage a limited dataset collected from a pilot study to learn an effective intervention policy for subsequent deployment. $\DPI$ was designed specifically for this setting. The intended usage of $\DPI$ therefore follows three steps. First, run a policy $\pi$ and let $\mathcal{D}$ be the dataset of trajectories of states, actions, and rewards. 
Second, use the dataset $\mathcal{D}$ to compute estimates $\hat{q}_{it}^{\pi}(S, A)$ \added{for ${q}_{it}^{\pi}(S, A)$} using a prediction algorithm. Third, deploy $\DPI(\pi)$ using the estimated intervention values, $\hat{z}^{\pi}_{it}(S) = \hat{q}_{it}^{\pi}(S, 1) - \hat{q}_{it}^{\pi}(S, 0)$.
Note that the main non-trivial step is the second one, the estimation. $\DPI$ was designed in a way to make this step as easy as possible. In particular, $\DPI$ can essentially be described by two design choices, both of which are motivated by the goal of easing estimation:
\begin{enumerate}
		\item[(a)] It performs \textit{one} round of policy iteration.
		\item[(b)] It uses $q$-values that are \textit{decomposed} at the patient level.
\end{enumerate}	
We explain both of these design choices below.

\subsubsection{Design choice (a): A single iteration.}
Given a dataset $\mathcal{D}$ generated from running $\pi$, estimating $q_{it}^\pi(S, A)$ corresponds to a \textit{prediction} problem (also called on-policy evaluation).
Prediction is one of the simplest tasks in reinforcement learning, 
and there exist a myriad of techniques that one can use for this task, such as Monte Carlo methods or temporal difference learning \citep{sutton2018reinforcement,szepesvari2022algorithms}.
Because of this significant existing literature on prediction, our work does not focus on analyzing this step.
In \cref{s.casestudy}, we provide an example of this step using a linear function approximation for the $q$ function.

However, to do \textit{more than one} round of policy iteration, one needs to learn $q^{\pi'}$ for a policy $\pi' \neq \pi$.
Estimating $q^{\pi'}$ from data generated from $\pi$ is an \textit{off-policy evaluation} problem, which is provably hard; \cite{wang2020statistical} show that the sample complexity of off-policy evaluation with a linear function approximation is \textit{exponential} in the time horizon.
Motivated by the hardness of off-policy evaluation, a recent work in the general offline RL literature, \cite{brandfonbrener2021offline}, also proposes the method of doing one-step of policy iteration from offline data and demonstrates strong empirical results of this approach.
Our work is complementary to the empirical findings of \cite{brandfonbrener2021offline},
\added{though we note that our policy is not the same as the one studied in \cite{brandfonbrener2021offline} due to the decomposition, discussed next.}

\subsubsection{Design choice (b): Decomposition.} \label{sss.designchoiceb}
The main difference between $\DPI$ and one iteration of the usual policy iteration (PI) is that $\DPI$ operates on the $q$-function, which is specific to an individual patient.
PI, in contrast, operates on the state space of the system MDP.
Specifically, for a state profile $\bS = (S_1, \dots, S_N)$ and action profile $\bA = (A_1, \dots, A_N)$, PI operates on the function $Q_{t}^\pi(\bS, \bA)$ defined as
\begin{align}  \label{eq:Q}
Q_{t}^\pi(\bS, \bA) = \bE_{\pi}\left[\sum_{t'=t}^T \sumN R(S_{it'}^\pi, S_{i,t'+1}^\pi, A_{it'}^\pi) \;\bigg|\; S_{it}^\pi = S_i, A_{it}^\pi = A_i \; \forall i \in [N] \right].
\end{align}

The advantage of the decomposition (using $q$ rather than $Q$) is in its ease of estimation.
Note that the size of the state and action space for the system MDP is $|\cS|^N$ and $N \choose B$, respectively, whereas the size of the state and action space for a single patient is $|\cS|$ and 2, respectively. Then, if one used a tabular approach to estimating these functions, learning $Q_t$ requires estimating $|\cS|^N {N \choose B}$ quantities, while learning $q_{it}$ for every $i$ requires estimating $2|\cS|N$ quantities. The latter is an \textit{exponentially} fewer number of estimation tasks than the former.

The disadvantage of the decomposition is that since the individual $q$ functions do not take as input the entire system state, it loses information. That is, for a system state $\bS =(S_1, \dots, S_N)$ and action profile $\bA =(A_1, \dots, A_N)$, the sum of the $q$-values do not correspond to the $Q$-value: $\sum_{i=1}^N q_{it}^{\pi}(S_i, A_i)\neq Q_t^{\pi}(\bS, \bA)$.
This introduces a bias in the improvement step of policy iteration.
This can lead to an undesirable behavior where policy iteration does not always result in an improvement \added{(formalized next in \cref{prop:dpi_can_be_worse})}.

Therefore, the decomposition improves the ease of estimation, but comes at a cost of losing information.
In \cref{s.theory}, where we study a stylized model, we establish a performance guarantee based on the decomposed $q$ values.
This guarantee relies on the baseline policy, $\pi$, to be randomized, and the performance guarantee improves as the number of interventions that $\pi$ gives decreases.
Both of these features, randomization and a small number of interventions, aids in reducing the extent of impact of the bias. 

\subsection{\added{Theoretical Guarantees for the Full Model}} \label{ss.general_guarantees}
\added{
Generically, vanilla policy iteration results in a guaranteed improvement in performance at each step \citep{sutton2018reinforcement}.
In our model, one step of policy iteration using the \emph{system} Q-values (\cref{eq:Q}), from policy $\pi$ would result in an improved policy over $\pi$.
However, Proposition \ref{prop:dpi_can_be_worse} shows that this is not always the case for $\DPI(\pi)$ due to the decomposition.

\begin{proposition} \label{prop:dpi_can_be_worse}
There exists an instance $\mathcal{I}$ and a policy $\pi$ where the expected total reward of $\DPI(\pi)$ is strictly smaller than that of $\pi$.
\end{proposition}

Although $\DPI(\pi)$ does not universally guarantee an improvement, the following result shows that this is the case when $\pi$ belongs to a particular class of policies that gives interventions in a random manner.
For $\gamma \in (0, 1)$, let $\RAND(\gamma)$ be the policy in which at every time step, gives an intervention to every patient independently with probability $\gamma$.
This class of policies is an approximation of the policy that was deployed for Keheala.

\begin{theorem} \label{thm:improvement_random}
Let $\gamma \in (0, 1)$ and suppose the budget $B$ satisfies $B \geq \gamma N$.
Then, for any instance, $\DPI(\RAND(\gamma)) \geq \RAND(\gamma)$.
\end{theorem}

While this result demonstrates that our policy results in an improvement if the baseline policy is $\RAND(\gamma)$, the improvement can be marginal, as \cref{thm:improvement_random} does not speak to the \emph{magnitude} of the improvement.
In the next section, we study a special case of MDPs that is inspired by our application, and we provide a different theoretical guarantee which is substantially stronger. 
}

\section{\added{Theoretical Guarantees for a Special Case}} \label{s.theory}
The full model, presented in the previous section, describes the intended use case for $\DPI$. 
However, the generality of this model prohibits the ability to have strong theoretical guarantees.
In this section, we therefore introduce a special case of the model that is inspired by our problem setting,
where we are surprisingly able to prove strong performance guarantees.
We first introduce this stylized model in \cref{ss.2statemodel} and describe the form of $\DPI$ in this setting in \cref{ss.simplepolicy}. We then present our performance guarantees in \cref{ss.perfguar} and additional robustness results in \cref{s.robustness}.

\subsection{Two-State MDP Model} \label{ss.2statemodel}
\begin{figure}[h]
\begin{center}
  \includegraphics[width=0.55\linewidth]{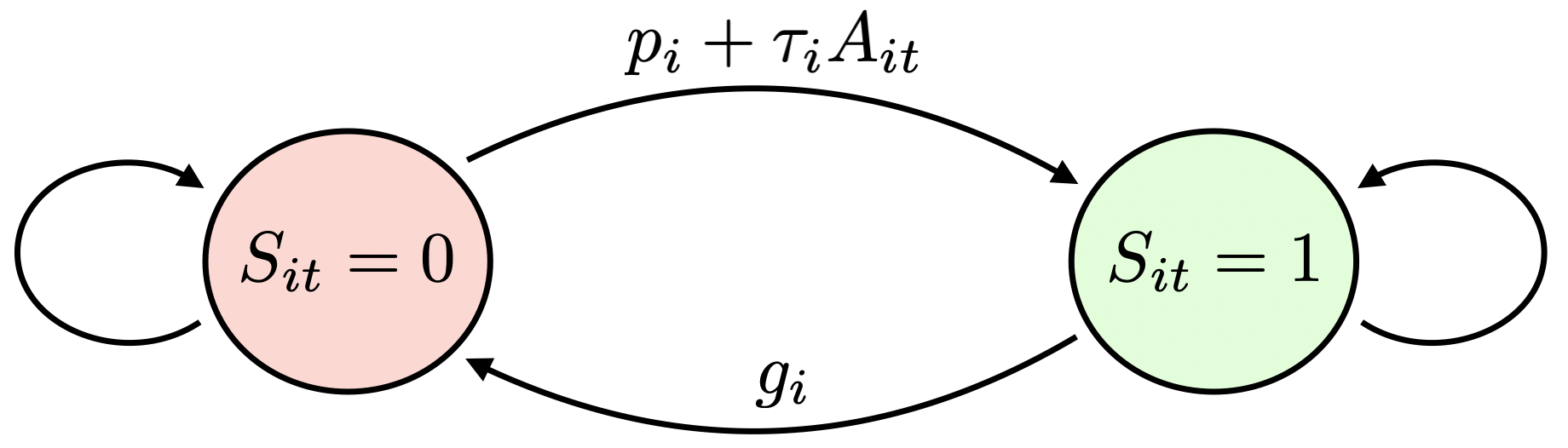}
  \caption{MDP for patient $i$.}
  \label{fig:2_state_mc}
\end{center}
\end{figure}

The state space is $\cS = \{0, 1\}$, where 0 and 1 correspond to an undesired and desired state, respectively.
Under the null action ($A=0$), $p_i$ and $g_i$ represent the probability of transitioning from state 0 to 1 and 1 to 0, respectively.
The intervention ($A=1$) only changes the probability of transitioning from state 0 to 1, which becomes $p_i + \tau_i$.
We assume that $p_i, g_i \in [0, 1/2]$ for all $i$, implying that states are more likely to stay the same than change when there is no intervention.
We assume $\tau_i \in [0, 1-p_i]$ for all $i$, hence an intervention can only increase the probability of switching from state 0 to 1. 
Lastly, the reward is simply equal to the resulting state; i.e.,\ $R_i(S, S', A) = S'$.
This represents the goal of maximizing the fraction of time that the patient is in the desired state.
The MDP for one patient is specified by the parameters $(p_i, g_i, \tau_i)$, and this MDP is summarized in Figure~\ref{fig:2_state_mc}.

This is a special case of the full model from \cref{s.gmodel}, where we assume the patient MDP takes the simple form described above.
All other aspects of the model remain the same. In particular, the system MDP is derived by combining the patient MDPs via a budget constraint. Note that the set of possible policies is finite, since both the state and action spaces are finite. Therefore, for any instance, there exists an optimal policy $\OPT$ which maximizes the objective, $\bE[\sumT \sumN S_{i,t+1}]$.

We believe this simplified version of the system is representative and relevant, both for our specific motivating application and behavioral health operations in general. From the perspective of Keheala (as described in \cref{s.intro}), the stylized model captures the salient features of patient behavior in the simplest way possible.
A dataset collected by an RCT run by Keheala revealed that the best single feature that predicted whether a patient will verify on a given day, is simply whether they verified the day before (80.9\% accuracy).
Therefore, the states in the model simply represent whether a patient verified the previous day.
\added{Additionally, according to the Keheala protocol, the subjects eligible for outreach interventions are those who have not been verifying treatment adherence, the objective being to encourage behavior change. Since the protocol does not prescribe outreach to subjects who are in the desired state (i.e., verifying their treatment adherence), our model is such that the intervention only impacts the transition from state 0 to state 1.}

More broadly, similar Markov models have been used in the literature to model patient behavior.
\cite{mate2020collapsing} and \cite{mate2022field} use the same two-state model and apply it to the setting of TB treatment adherence and maternal health respectively. \cite{biswas2021learn} also studies maternal health in which they employ a similar MDP with three states. 
These three works also use the same objective function as we do.
This type of a two-state Markov model is originally from the literature on communication systems (where it is referred to as a Gilbert-Elliot channel \citep{gilbert1960capacity}) but has also been used in marketing \citep{schmittlein1987counting}.

\subsection{The $\DPI$ Policy for the Two-State MDP Model} \label{ss.simplepolicy}
Our main result pertains to $\DPI(\pi)$ for $\pi$ belonging to a specific class of policies.
Specifically, for $\gamma \in (0, 1)$, let $\RAND(\gamma)$ be the policy in which at every time step, gives an intervention to every patient in state 0 independently with probability $\gamma$.
Denote the intervention values of the policy $\RAND(\gamma)$ by $z^\gamma_{it} \triangleq z^{\RAND(\gamma)}_{it}$.
We will provide a performance guarantee for the policy $\DPI(\gamma) \triangleq \DPI(\RAND(\gamma))$.

Our focus is in the regime where $\gamma$ is small (close to 0), and we will see that the theoretical guarantees are stronger when this is the case.
This represents a `budget-constrained' regime; the probability of a patient receiving an intervention is small, or equivalently, the number of interventions given is small relative to the number of patients.
To give intuition on the policy $\DPI(\gamma)$ in this regime, we provide a simple expression for the intervention value in the case where $\gamma \to 0$ and $T \to \infty$ (the latter limit is only assumed to provide a simple expression).
\begin{proposition} \label{prop:z_clean_form}
Fix a patient $i$ and time step $t$.
Then, $\lim_{T \to \infty} \lim_{\gamma \to 0^+} z^{\gamma}_{it}(0) = \tau_i/(p_i + g_i)$.
\end{proposition}

\added{This result is proven as part of \cref{lemma:zgamma} in Appendix~\ref{sec:coupling}.}
\cref{prop:z_clean_form} says that when $T \to \infty$ and $\gamma \to 0$, $\DPI(\gamma)$ orders patients in state 0 by the index $\tau_i/(p_i + g_i)$, and intervenes on those with the highest values.
The index increases with $\tau_i$ and decreases with $p_i$ and $g_i$, which is intuitive. 
$\tau_i$ is the `treatment effect' of an intervention on the probability that the patient switches from state 0 to 1.
The smaller the value of $g_i$, the longer the patient will stay in state 1 once they are there, increasing the `bang-for-the-buck' of an intervention.
As for $p_i$, a larger value means that a patient is likely to move to state 1 \textit{without} an intervention; hence lower priority is given to those with a large $p_i$.

\subsection{Performance Guarantee}\label{ss.perfguar}
Let $\NULL$ be the policy that only takes the null action; i.e.\ $\NULL_t(\bS) = (0, \dots, 0)$ for all $\bS, t$.
As a slight abuse of notation, we use $\OPT, \NULL, \RAND(\gamma)$ and $\DPI(\gamma)$ to also denote the expected reward of those policies.
We now state our main result, a performance guarantee for the quantity $\DPI(\gamma) - \NULL$, compared to $\OPT - \NULL$.
\begin{theorem}[Main Result] \label{thm:approx_ratio_b}
Given an instance of the two-state model, let $\bar{M} = \max_{i \in [N]} \frac{\tau_i (1-p_i-g_i)}{(p_i + g_i)(1-p_i)} > 0$. For any $\gamma \in (0, 1)$,
\begin{align} \label{eq:approx_ratio_b}
\DPI(\gamma)  - \NULL \;\geq\; \frac{1}{2(1+\gamma \bar{M})} \; ( \OPT - \NULL ).
\end{align}	
\end{theorem}

\cref{thm:approx_ratio_b} provides an approximation guarantee for $\DPI(\gamma)$ with respect to the \textit{improvement} over the null policy.
The approximation ratio, $\frac{1}{2(1+\gamma \bar{M})}$ depends on two quantities: $\gamma$ is a parameter of the policy $\RAND(\gamma)$ which is used to compute the intervention values, and $\bar{M}$ is a quantity related to the instance parameters, $\{(p_i, g_i, \tau_i)\}_{i \in [N]}$.
For any fixed instance, the approximation ratio improves as $\gamma$ decreases, approaching 1/2 as $\gamma$ goes to 0.
Therefore, as $\gamma \to 0$, $\DPI(\gamma)$ achieves at least half of the improvement in reward as compared to that of the optimal policy.

With respect to $\bar{M}$, a smaller value implies a stronger guarantee.
Since we assume $p_i, g_i \leq 1/2$, $\bar{M} \leq \max_{i \in [N]} \frac{\tau_i}{p_i + g_i}$.
For $\bar{M}$ to be small, it must be that $p_i + g_i$ is bounded away from 0 for all $i$ --- specifically, patients must not be \textit{completely} sticky, where sticky means that they never switch their state, no matter which state they are in.

\cref{thm:approx_ratio_b} is significantly stronger than and implies the \textit{usual} notion of an approximation result, which is the following:
\begin{corollary}[Weaker result]  \label{thm:weaker_result}
For any problem instance of the two-state model,
$\DPI(\gamma) \geq  \frac{1}{2(1+\gamma \bar{M})} \OPT$.
\end{corollary}
Because a patient can transition from state 0 to 1 \textit{without} an intervention (if $p_i > 0$), the $\NULL$ policy can achieve a significant fraction of $\OPT$.
If it was the case that $\NULL$ is more than half of $\OPT$, then \cref{thm:weaker_result} would be vacuous.
On the other extreme, if an intervention was \textit{necessary} for all patients to transition to state 1
(i.e.\ $p_i = 0$ for all $i$), then it would be that $\NULL = 0$, in which case the result of \cref{thm:approx_ratio_b} is equivalent to that of \cref{thm:weaker_result}.

\added{We make three observations about our main result. First, the complete proof of \cref{thm:approx_ratio_b} can be found in Appendix~\ref{s.app.proof}, but to provide intuition we note that it relies on a key connection of our model to the problem of online allocation of reusable resources. In that problem, a set of items are sold to arriving customers over time, and the items are `reusable' in that the customer returns the item after some period of time.	One can interpret our problem as one of reusable resources, in which patients are analogous to `items' and the item is `available to sell' if the patient is in state 0, and the item has `been sold' if the patient is in state 1. Giving an intervention to a patient can be thought of as `offering the item in an assortment', in which the item will be bought with some probability. For \cref{thm:approx_ratio_b}, we leverage a sample path coupling proof technique of \cite{gong2021online}, who show a 1/2-approximation for the greedy policy for reusable resources. A direct application of this technique results in the weaker approximation result of \cref{thm:weaker_result}. \cref{thm:approx_ratio_b} requires developing a more intricate sample path coupling across the policies $\NULL$, $\OPT$, and $\DPI(\gamma)$.

Second, we believe the approximation guarantees can potentially be made tighter, although we leave that as an open direction. The reason for this is that the inspiration for our proof technique (the coupling idea used by \cite{gong2021online}) was developed for an \emph{adversarial} online allocation problem. In our setting such an assumption would imply that the set of patients who are in state 0 are adversarially chosen at each round. This is not the case since our model is stochastic, not adversarial, but our proof does not leverage this stochastic structure, and hence the bound may be able to be improved.

Third, with the question of tightness in mind, we run synthetic simulations with the two-state model to empirically evaluate the performance of $\DPI$, which we compare to both a random benchmark and the Whittle's index (which can be exactly computed due to the simplicity of the model). The results are included in Appendix~\ref{s.app.synthetic simulations}. Although the Whittle's index is not necessarily optimal (due to heterogeneous arms and the finite time horizon), it serves as a strong benchmark.
We observe that $\DPI$ greatly outperforms the random policy, and it performs almost identically to the Whittle's index.
}

\subsection{Robustness under an Approximate Implementation} \label{s.robustness}
Note that \cref{thm:approx_ratio_b} assumes that the intervention values are known exactly.
In practice, these intervention values must be estimated, as discussed in the intended usage in \cref{s.intended_usage}.
We show that this result is robust, in the sense that a policy that \textit{approximately} implements $\DPI(\lambda)$ also yields a performance guarantee.

Suppose $\ALG$ is an \textit{index policy} using indices $z^{\ALG}_{it}(0)$.
That is, $\ALG$ assigns interventions to the patients in state 0 with the largest value of $z^{\ALG}_{it}(0)$.
Then, we show that if the index values $z^{\ALG}_{it}(0)$ approximate the intervention values $z^{\lambda}_{it}(0)$, then $\ALG$ also admits a performance guarantee.

\begin{theorem} \label{corr:approx}
Suppose $\ALG$ is an index policy that uses index values $z^{\ALG}_{it}(0)$ that satisfies, for all $i$ and $t$,
\begin{align} \label{eq.approxvalues2}
c_1 z_{it}^{\lambda}(0) \leq z^{\ALG}_{it}(0) \leq c_2 z_{it}^{\lambda}(0),
\end{align}
for some $c_1 \leq 1$ and $c_2 \geq 1$.
Then,
\begin{align*}
\ALG - \NULL \;\geq\; \frac{c_1}{c_2} \cdot \frac{1}{2(1+\gamma \bar{M})} \; ( \OPT - \NULL ).
\end{align*}
\end{theorem}

The proof can be found in Appendix~\ref{sec:app:pf_approx}.
This result implies that one does not have to run $\DPI$ \textit{exactly} in order to achieve good performance. 
In practice, one may estimate the intervention values from data --- even if the estimation is not perfect, \cref{corr:approx} ensures a performance guarantee.

\section{Case Study: TB Treatment Adherence}\label{s.casestudy}
In this section, we revisit the full problem (introduced in \cref{s.model}) and conduct numerical experiments to evaluate the performance of $\DPI$ for its intended use case. Our analysis is motivated by our partner organization, Keheala, which operates a digital health platform to support medication adherence among TB patients in highly resource-constraint settings. Here, we first summarize the state of the global TB epidemic and the Keheala behavioral intervention (\cref{ss.keheala}). We then describe our data sources and the validated simulation model we have developed to test outreach policies (\cref{ss.data} and \cref{ss.sim}). Next, we discuss how our policy, as well as some benchmark policies, can be implemented using Keheala's data (\cref{ss.policies}), before presenting our numerical results (\cref{ss.results}).

\subsection{The global TB epidemic and the Keheala intervention}\label{ss.keheala}
TB remains one of the deadliest communicable diseases in the world, causing 1.6 million deaths in 2021. This is remarkable in light of the fact that effective treatment has been available for over 80 years, with the current WHO guidelines recommending a 6 month regimen of antibiotics for drug-susceptible TB and a more intense regiment for drug-resistant strains \citep{World22Global}. A key limiting factor for curbing the epidemic is lack of patient adherence to these treatment regiments, which increases the probability of infection spreading, drug resistance, and poor health outcomes \citep{garfein2019synchronous}. 

Keheala was designed to provide treatment adherence support to TB patients in resource-limited settings. Their platform operates on the Unstructured Supplementary Service Data (USSD) mobile phone protocol, which importantly allows phones without smart capabilities to access the service. Once a patient has enrolled with Keheala, they are meant to self-verify treatment adherence every day, using their mobile phone. In addition, they have access to a range of services. Some are on-demand, for example educational material about TB or leaderboards for verification rates. Others are automatic, such as adherence reminders, which are sent to patients daily (at their pre-determined medication time) in the absence of verification. In addition, the Keheala protocol is to escalate outreach interventions when patients do not self-verify adherence. It states that after one day of non-adherence patients should receive a customized message to encourage resumed adherence and after two days of non-adherence patients should receive a phone call from a support sponsor. While these support sponsors are full-time employees, they are not healthcare professionals. They are members of the local community who have experience with TB treatment and are therefore familiar with the many contributing factors associated with low treatment adherence, such as side-effects, societal stigma against TB patients, and challenges with refilling prescriptions.

The overall effectiveness of Keheala's combination of services was evaluated in a randomized controlled trial (RCT) in Nairobi, Kenya. The trial demonstrated that Keheala reduced unsuccessful TB treatment outcomes—a composite of loss to follow-up, treatment failure, and death—by roughly two-thirds, as compared to a control group that received the standard of care \citep{Yoeli19Digital}. Given this success, Keheala's primary practical objective is to ensure that enrolled patients remain engaged with the platform through adherence verification.

In this paper, we focus on the final level in Keheala's escalation protocol---support sponsors making phone calls to patients. This part of the outreach was organized through populating a daily list of patients who had not verified treatment adherence for 48 hours. Support sponsors had many responsibilities in operating the platform, but would make phone calls to as many patients on the list as possible on a given day. Since hiring support sponsors is a costly aspect of operating the service, Keheala is interested in implementing a more personalized and targeted approach for prioritizing which patients should receive a phone call on a given day. Being able to maintain a similar performance with fewer support sponsors (or equivalently, serve more patients with the same number of support sponsors) is desirable for any future scale-up of the system.

\subsection{Data sources.}\label{ss.data}
Based on the success of the first RCT, the effectiveness of Keheala was further evaluated in a second RCT\footnote{The trial was approved by the institutional review board of Kenyatta National Hospital and the University of Nairobi. Trial participants or their parents or guardians provided written informed consent. The trial’s protocol and statistical analysis plan were registered in advance with ClinicalTrials.gov (\#NCT04119375).} during 2018-2020. The RCT was conducted in partnership with 902 health clinics distributed across each of Kenya's eight regions, representing a mix of rural and urban clinics. The study included four treatment arms and enrolled over 15,000 patients. We obtained data for 5,433 patients enrolled in the Keheala intervention arm (other arms aimed to independently test specific components of the Keheala intervention). 

As part of the RCT, the study team collected socio-demographic information from all patients. This information includes static covariates such as age, gender, language preferences, location, as well as limited clinical history (see \cref{s.app.list_features} for a full list). In addition, Keheala collected engagement data about each patient during their enrollment in the service. This includes whether a patient verified on a given day, how many reminders they received, and whether they were contacted by a support sponsor. 

After filtering out patients with missing information or not enough data, we conducted all our analysis on 3594 patients. The average patient was enrolled on the platform for 118 days. On an average day, 608 patients were enrolled and 210 of those were eligible for a support sponsor call according to the protocol (i.e., having not verified treatment adherence for the preceding 48 hours). The support sponsors, employed by Keheala, had a range of responsibilities in operating the platform, including making outreach phone calls to the eligible patients. 
The average number of calls made per day was 25.5. Hence, in our analysis, we use a budget of $B$ = 26 as our main point of comparison.

\subsection{Simulation Model.}\label{ss.sim}
We build a simulation model that we use to estimate the counterfactual outcomes of different outreach approaches.
The simulator is effectively represented by a single function, $f(S, A) \in [0, 1]$, which denotes the probability that a patient in state $S$ with action $A$ verifies in the next time step.
This function is used to simulate one step transitions for every patient.
We first describe the state space of the patients, describe the exact simulation procedure, and then discuss how we learn $f$ from data and validate the simulator.

\subsubsection{Patient state space.}\label{ss.simstatespace}
For patient $i$, let $X_i \in \bR^{13}$ be their static covariates. 
Let $V_{it} \in \{0, 1\}$ denote whether patient $i$ verified at time $t$, and let $A_{it}\in \{0, 1\}$ denote whether the patient received the intervention at time $t$.
Let $H_{it} = (V_{i1}, A_{i1}, \dots, V_{i, t-1}, A_{i,t-1}, V_{it}) \in \bR^{2t-1}$ be the history of verifications and interventions up to time $t$.
We define a \textit{condensed} history $\tH_{it} \in \bR^{21}$ by summarizing the history $H_{it}$ into 21 features, aiming to capture as much relevant information as possible.
The condensed history contains the patient's recent and overall behavior.
For statistics such as the number of times the patient verified and the number of interventions they have received, we aggregate them over the past week, as well as in total.
We also include information on their verification and non-verification streaks, as well as how long they have been in the platform.
See \cref{s.app.list_features} for a full list of these features.
Then, we define the state of patient $i$ at time $t$ to be $S_{it} = (X_i, \tH_{it}) \in \bR^{34}$.

\subsubsection{Simulation procedure.} 
Given $f$ and an intervention policy $\pi$, we `mimic' the RCT by simulating patient behavior day by day. 
In total, we simulate $T=700$ time steps, where each $t \in [T]$ corresponds to one day between April 2018 to March 2020. We let $T_{s}(i)$ and $T_{e}(i)$ denote the starting and ending time steps that patient $i$ was enrolled in Keheala. Each patient $i$ is then introduced into the system at time $t=T_s(i)$, and removed at time $t=T_e(i)$. We use the observed data from the RCT for their first 7 days in the system to initialize their state. 
Then, in each time period, given the set of patients that were active in the RCT for more than 7 days, we use a policy $\pi$ on these patients to determine who receives sponsor outreach. If a patient $i$ was in state $S_{it}$ at time $t$ and the policy $\pi$ chose action $A_{it}$, we let $V_{i, t+1}$ be 1 with probability $f(S_{it}, A_{it})$, and 0 otherwise (where the randomness is independent across patients and time steps).
Finally, we use $V_{i, t+1}$ to update their state for the next time step. 

\subsubsection{Estimating the $f$ function.} \label{s.learnf} 
Using the state space $\cS$ as described above, we construct the function $f: \cS \times \{0, 1\} \to [0, 1]$ using data from the RCT. Specifically, we learn the two functions $f(S, 0)$ and $f(S, 1)$ separately. For $f(S, 0)$, we train a gradient boosting classifier on the dataset $\{(S_{it}, V_{i,t+1})\}_{i \in N, t \in [T] : A_{it} = 0}$, using $V_{it}$ as the outcome variable. For $f(S, 1)$, we write the function as $f(S, 1) = f(S, 0) + \tau(S)$ and we learn $\tau(S)$ using the double machine learning method of estimating heterogeneous treatment effects \citep{chernozhukov2018double}. See \cref{s.app.simulation} for details on implementing this method.

\subsubsection{Train and test split.} \label{s.traintestsplit}
Importantly, we use a different set of patients to estimate the $f$ function (and to run our simulations) from the set of patients we use to train our policies. In particular, we randomly split all patients from the RCT into two groups, which we call \textit{train} and \textit{test}. We use the \textit{test} set to estimate the $f$ function that forms the basis for the simulation model. We keep the \textit{train} set of patients separate and use it to train policies (see \cref{ss.policies}). This ensures that the policies we evaluate are not learned off of the same dataset that was used to learn the simulator. 
The simulation itself uses the test patients, and we duplicated each patient so that we maintain a similar total number of patients as in the original study.

\subsubsection{Simulation validation.}
We validate the performance of the simulator on a \textit{different} intervention policy than the simulator was trained on, by leveraging the fact that there was variability in the number of interventions given throughout the RCT.
In particular, the average number of interventions given during the first half of the RCT was around double of that of the latter half (45.9 vs. 21.4),
and this variation induces a change in the intervention assignment policy.
Then, dividing the data into halves produces two datasets that are generated using effectively different intervention policies.

To validate the simulator, we use the method from \cref{s.learnf} to learn $f$ using the first dataset, and then validate its performance on the second dataset. 
Using this procedure, the AUCs on the second dataset for $f(S, 0)$ and $f(S, 1)$ were 0.918 and 0.745, respectively.
We also check the calibration of both of these functions, by grouping the samples into bins based on their predicted probability of verifying the next day, and checking whether their actual verification rates.

\begin{figure}[h]%
	\centering
	\subfloat[Calibration for $f(S, 0)$]{{\includegraphics[width=0.48\linewidth]{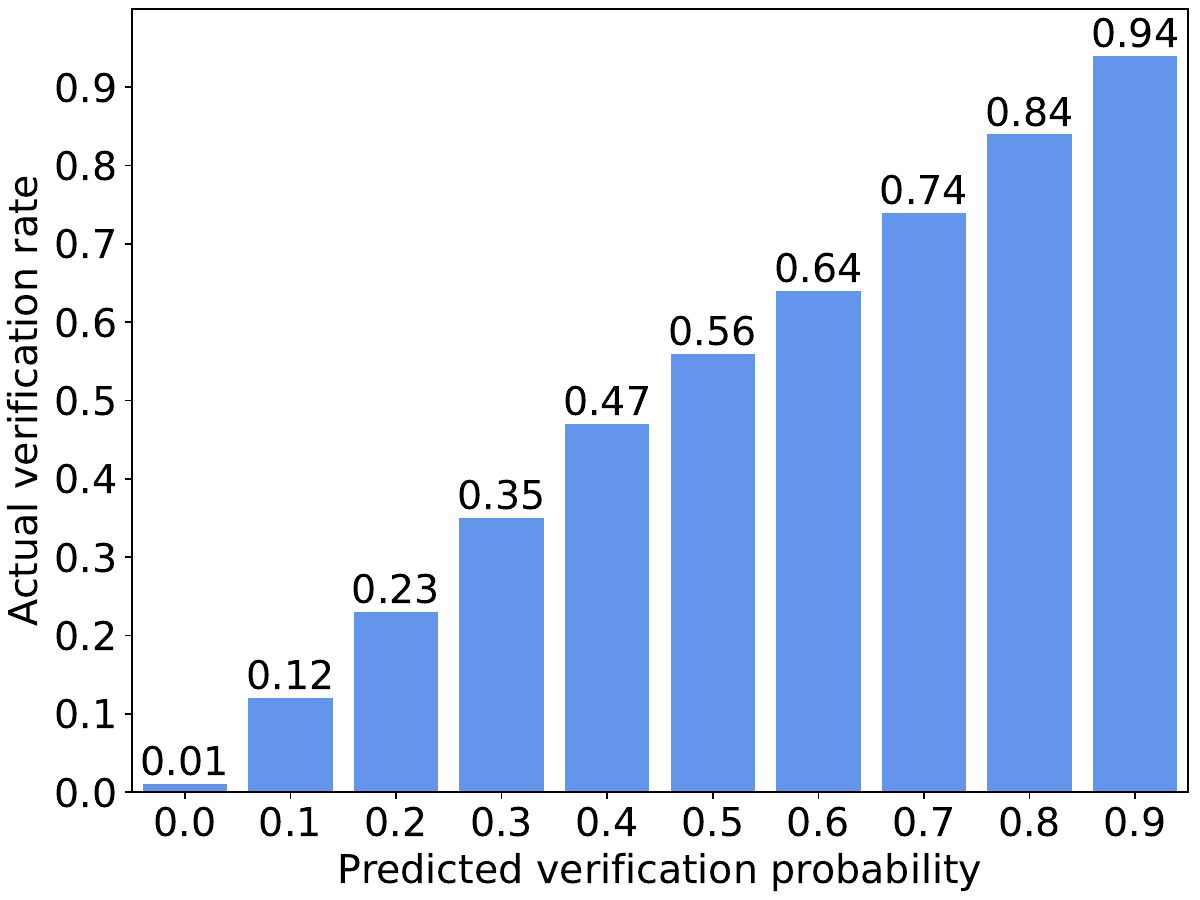} }}%
	\subfloat[Calibration for $f(S, 1)$]{{\includegraphics[width=0.48\linewidth]{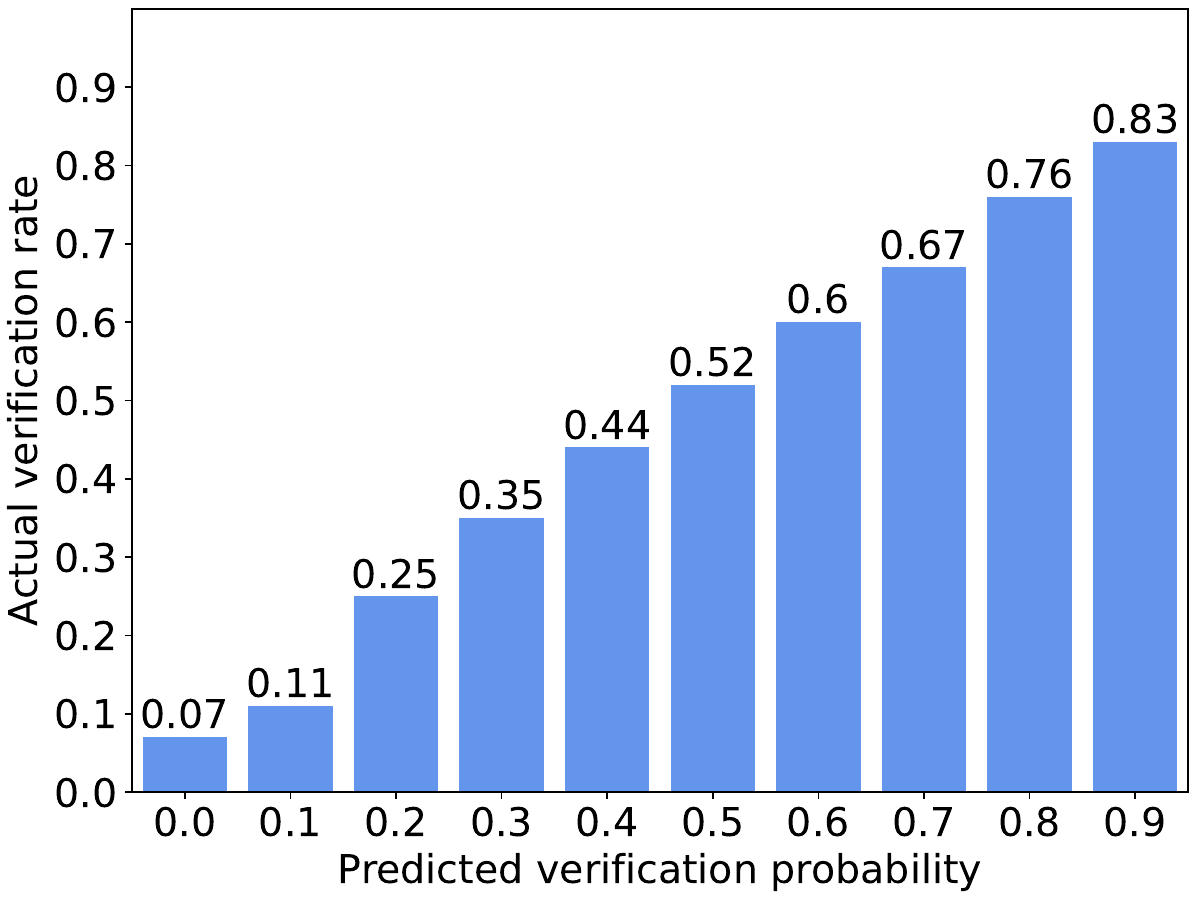} }}%
	\vspace{2mm}
	\caption{Calibration plots for $f(S, 0)$ and $f(S, 1)$ for simulation validation. 
		We group the samples based on the simulated probability of verification
		into bins with a 10\% range, which we label by the lower number. 
		For example, the 0.3 bin on the x-axis represents the samples whose probability of verification according to $f$ is in $[0.3, 0.4)$; hence we should expect the actual number of verifications of those samples to be close to 0.35.}
	\label{f.calibration}%
	\vspace{-2mm}
\end{figure}

We group the samples based on the simulated probability of verification into bins with a 10\% range, and we compute the expected calibration error (ECE) \citep{naeini2015obtaining}. For bin $i$, let $o_i$ be the true fraction of positive instances in bin $i$, $e_i$ be the mean of the predicted probabilities of the instances in bin $i$, and $N_i$ be the number of samples in bin $i$. Then, the ECE is defined as
\begin{align}
\text{ECE}	 = \frac{1}{N} \sum_{i=1}^{10} N_i |o_i - e_i|,
\end{align}
where $N$ is the total number of samples.
The expected calibration error was 0.0066 for $f(S, 0)$ and 0.0308 for $f(S, 1)$.
Figure~\ref{f.calibration} displays these bins.

These results demonstrate that the simulator has good performance in mimicking patient behavior. 
As expected, the AUC and the ECE is worse for $f(S, 1)$ compared to $f(S, 0)$; 
this is due to the \textit{significantly} fewer number samples with an intervention in the training data, as well as the increase in variance of doing off-policy estimation.
The training data used for $f(S, 0)$ had 300K samples, while the one used for $f(S, 1)$ had 4.5K samples.
For $f(S, 1)$, the calibration is slightly off for samples with a high probability of verification (bins 0.7-0.9); however, we note that the 0.7-0.9 bins only contain 11.3\% of all samples for $f(S, 1)$.

\subsection{Outreach Policies and Experimental Design}\label{ss.policies}
Using the simulation model described above, we compare the performance of three main policies. For each policy, we vary the budget for outreach interventions per day between 10 and 40. As mentioned before, the average number of sponsor outreaches during a given day of the trial was 26. Importantly, we restrict all policies so that they can only provide an outreach to patients who have not verified for at least two days in a row. This is because that was what was done in the RCT, hence there is no data for how an outreach affects behaviors for those who do not meet this criterion (thus we would not be able to accurately evaluate policies that do not follow this restriction).

We note that attaining improved performance with lower outreach capacity is particularly important for the resource-limited setting at hand as it speaks to the performance achievable during a future scale-up of the system, in which the ratio of patients to support sponsors is likely to be much higher. 
Next, we describe the implemented policies.

\subsubsection{$\DPI$ for Keheala.} 
The first step in operationalizing $\DPI$ is defining the state space for each patient. For this, we use the same condensed state space as described in \cref{ss.simstatespace}, i.e., we define the state of patient $i$ at time $t$ to be $S_{it} = (X_i, \tH_{it}) \in \bR^{34}$ (a full list of these features is included in \ref{s.app.list_features}). Importantly, we note that all of the features of this state space are observable to Keheala at any time $t$, once a patient has been enrolled on the platform for seven days. 

The second step is estimating the $\hz_{it}(S_{it})$ score for each patient at each time period, which is ultimately used to prioritize patients. 
As before, we let $T_{s}(i)$ and $T_{e}(i)$ be the starting and ending time steps that patient $i$ was enrolled in Keheala. Using this notation, we can represent the future verification \emph{rate} for patient $i$ at time $t$ by $y_{it} = \frac{1}{T_{\text{e}}(i)-t} \sum_{r=t+1}^T V_{ir}$.
Then, the data from the RCT can be written in the form $\{(S_{it}, A_{it}, y_{it})\}_{i \in [N], t \in \{T_{s}(i), \dots, T_{e}(i)\}}$, and we can estimate the function $q_{it}^{\baseline}(S, A)$ using this data.
In our implementation, we use a linear function approximation for the verification rate, assuming the form 
\begin{align*}
q_{it}^{\baseline}(S, A) = \langle \theta_A, S \rangle \cdot (T_{\text{e}}(i)-t),
\end{align*}
for each of the two actions $A \in \{0, 1\}$.
The $\langle \theta_A, S \rangle$ term represents the future verification rate, and $T_{\text{e}}(i)-t$ represents the number of days left; combined, $q_{it}^{\baseline}(S, A)$ represents the total number of future verifications.
We note that the state contains information regarding the number of days the patient has been enrolled in Keheala, hence the verification rate is also a function of the time step.

We estimate $\theta_a$ using least squares with an $\ell_2$ regularizer:
\begin{align} \label{eq:leastsquares}
	\hat{\theta}_a &\in \argmin_{\theta \in \bR^{34}} \bigg( \sum_{i \in N} \sum_{t=T_{\text{s}}(i)}^{T_{\text{e}}(i)}  \bI(A_{it} = A)(y_{it} - \theta^\top S_{it})^2 + ||\theta||^2_2 \bigg).
\end{align}

Finally, we compute a patient's estimate of their intervention value at time $t$ as
\begin{align*}
	\hz_{it}(S_{it}) = \langle \htheta_1 - \htheta_0,S_{it} \rangle \cdot (T_{\text{e}}(i)-t), 
\end{align*}
and the resulting policy is to give the intervention to up to $B$ patients with the highest positive $\hz_{it}(S_{it})$ values.

\subsubsection{Bandit.}
The bandit policy aims to choose patients with the highest increase in the probability of next-day verification, using a linear contextual bandit model. In terms of the two-state model from \cref{ss.2statemodel}, the goal is to choose patients with the highest value of $\tau$.
We essentially use the same state space and linear model as was used for $\DPI$, except that the outcome variable is next-day verification, rather than total future verifications.
We first learn a prior using the offline data, and then we run a Thompson sampling style policy, which continually updates the policy with online data.

Specifically, we assume the linear form $V_{i,t+1} = \langle \beta_a, S_{it} \rangle$ for action $a \in \{0, 1\}$, with unknown parameters $\beta_0, \beta_1 \in \bR^{34}$.
The prior on $(\beta_0, \beta_1)$ is initialized as the output of a least-squares regression using the offline data, the same data that was used to train $\DPI$.
At each time step, $(\tilde{\beta_0}, \tilde{\beta_1})$ is sampled from the posterior. 
Then, the policy chooses the $B$ patients with the highest value of $\langle \tilde{\beta_1}, S_{it} \rangle - \langle \tilde{\beta_0}, S_{it} \rangle$.
After the outcome is observed at each time step, the posterior is updated accordingly.
The detailed description on the algorithm can be found in Section~\ref{sec:app:bandit}.

This policy makes use of strictly more data than $\DPI$, since $\DPI$ only uses the offline data. 
In the results, we confirm that this policy indeed learns myopic rewards correctly.
Therefore, this is a very strong benchmark algorithm for optimizing myopic rewards.

\subsubsection{Whittle's index (QWI).} \label{sec:qwi_description}
The next benchmark is the Whittle's index.
The advantage of this method compared to the bandit benchmark is that is is non-myopic.
However, the downside is that computing the Whittle's index requires the model to be known.
To implement Whittle's index in our setting where the model is unknown, we leverage the recent work of \cite{avrachenkov2022whittle} who propose a Q-learning approach to learn the Whittle's index, which we refer to as $\QWI$.
$\QWI$ is an online learning method that simultaneously learns the Q-values as well as the Whittle's index for each state.

There are two main challenges in implementing $\QWI$ in our setting. The first is that the algorithm is an online learning method, and the second is that it requires a finite state space as it learns the Whittle's index for each state separately. 
For the first point, we adapt the algorithm from \cite{avrachenkov2022whittle} to an offline setting so that it can use the same data that is used to train $\DPI$.
For the second point, we define a smaller, finite state space so that $\QWI$ can be implemented.
We define a patient's state at a point in time to be a 3-tuple $(s_1, s_2, s_3)$, where $s_1$ represents the number of times the patient verified in the last week, $s_2$ is the patient's historical total verification rate, and $s_3$ is the number of times that the patient received an intervention in the last week. The values of each of these terms are bucketed into a small number of bins (3 bins for $s_1$ and $s_3$, 5 bins for $s_2$), resulting in 45 states in total. Specifically, the bins for both $s_1$ and $s_2$ are $0, 1$ and $2-7$. For $s_2$, the bins are $0-1\%$, $1-5\%$, $5-20\%$, $20-45\%$, and $45-100\%$. The bin values were chosen to balance the number of samples in each bin. 
This results in 45 states in total.

Based on this state space, we learn the Whittle's index, $\lambda(s) \in \bR$, for each state $s$.
Then, at each point in time, $\QWI$ chooses the patients in states with the highest Whittle's index to give the intervention to.
Further details of the learning algorithm is deferred to Appendix~\ref{sec:app:qwi}. 

\added{
We note that the state space for $\QWI$ is different from that of $\DPI$, due to the computational limitation of $\QWI$. In \cref{ss.state_space_results}, we run simulations where we modify the state space for $\DPI$ to be the same as $\QWI$, so that we can isolate the performance difference to the algorithm rather than the state space.
That said, we believe that the ease of working with an infinite state space is a substantial advantage of $\DPI$.
}

\subsubsection{Baseline.}
The $\baseline$ policy approximates the heuristic followed by Keheala in the two RCTs that have been implemented. In both cases, the protocol was that patients were added to the support sponsor call queue after not verifying for 48 hours. As a result, the order of patients in the queue is effectively random, determined by a combination of their designated medication time (which prompts automated reminders to take the medicine and verify) and the timing of their self-verification. We approximate the resulting outreach policy by selecting $B$ patients out of all those who have not verified for 48 hours, at random. 

\added{
The $\RAND$ policy used in the theoretical results is aimed to be an approximation of $\baseline$. 
The discrepancy between these policies is solely for technical convenience, 
as $\RAND$ is easier to analyze due to the independence across patients.
}

\subsubsection{Null policy.}
\added{
Lastly, we simulate the $\NULL$ policy which does not give any interventions. 
We note that this policy does not depend on the budget parameter $B$.
}

\subsection{Results}\label{ss.results}

\begin{figure}[h]
\begin{center}
\vspace{-3mm}
  \includegraphics[width=1\linewidth]{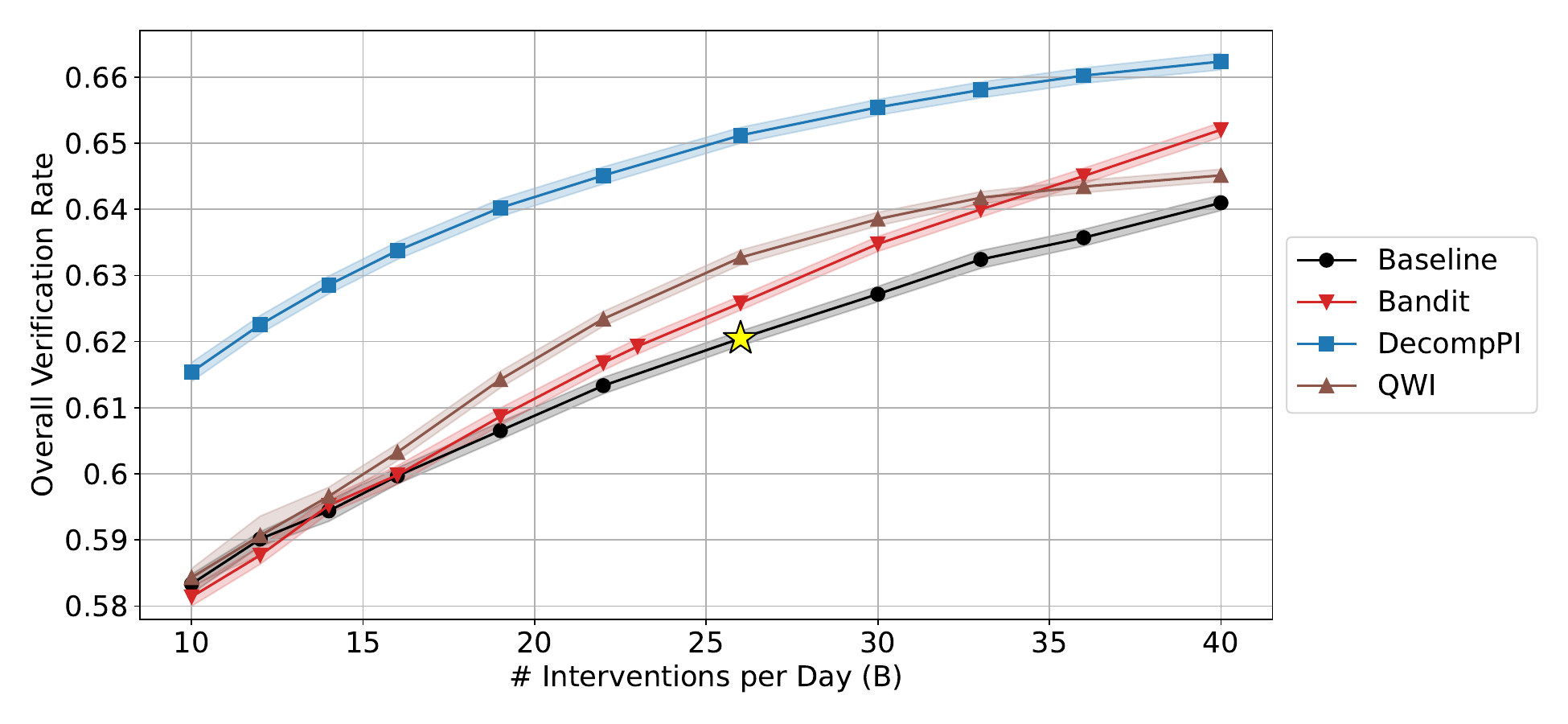}
  \caption{Average overall verification rate over 50 runs for each policy and budget. 
  The overall verification rate for the $\NULL$ policy was 54.2\%.
  The shaded region indicates a 95\% confidence interval. The star represents the operating point for Keheala.}
  \label{fig:main_results}
\vspace{-6mm}
\end{center}
\end{figure}

The results are shown in Figure~\ref{fig:main_results}. 

\subsubsection{Overall performance.}
The average performance for each budget and policy over 50 runs are shown in Figure~\ref{fig:main_results}, which shows that $\DPI$ clearly outperforms the other policies over a wide range of budget values.
For a practical interpretation of the results, consider \textsf{Baseline} at a budget of 26, the policy and budget that Keheala was operating during the RCT, which results in an overall verification rate of 62.0\%.
By using less than \textit{half} of the budget, $B=12$, $\DPI$ achieves the same verification rate at 62.2\%.
As the costliest aspect of Keheala's system is in hiring staff to provide the interventions, these results imply that they can cut these costs by half to achieve the same outcome.
\added{
The $\NULL$ policy (no interventions) results in a verification rate of 54.2\% (we did not plot this for readability of the figure).
One can interpret this number as a reference benchmark to compare the effectiveness of interventions.
When the budget is 26, $\baseline$ improves over $\NULL$ by 14.6\%, while $\DPI$ improves over $\NULL$ by 20.3\%. 
Therefore, $\DPI$ improves the effectiveness of the interventions over $\baseline$ by 38.3\%.
}

Additionally, we observe that the improvement of $\DPI$ compared to the other policies is especially substantial for smaller budgets. 
This implies that when the number of patients that can be targeted is small, $\DPI$ can correctly identify the set of patients to target that result in the largest gains.
This is especially valuable for scaling up the system.
Indeed, if Keheala wanted to expand to include more patients without linearly increasing their staff costs, then the ratio of budget to the number of patients would decrease, resulting in the regime where $\DPI$ offers major improvements.

The fact that the performance of $\bandit$ policy improves over $\baseline$ as the budget increases is caused by the increase in relevant data.
Note that the number of interventions is small ($\sim 26$) relative to the number of patients in the system at once ($\sim 600$), implying that the number of data points with $A=1$ is much smaller than that of $A=0$. 
Therefore, the main bottleneck in estimation is learning patient behaviors after receiving an intervention.
As the budget increases, the $\bandit$ has access to more data from patients with an intervention, and hence is able to improve its learning. 

$\QWI$ has a slightly inconsistent performance curve relative to the other policies.
Its performance is always better than or similar to $\baseline$, but compared to $\bandit$, it over-performs in the mid-budget regime, but under-performs as the budget increases.
We dive deeper into the types of patients each policy targets in \cref{sss.targetedpatients}, where we provide an explanation for this behavior.

\added{
One factor that may be contributing to the poor performance of $\QWI$ is the state space that is used.
$\QWI$ uses a discretized state space as described in \cref{sec:qwi_description}, different than the infinitely-sized state space used by $\DPI$.
In \cref{ss.state_space_results}, we run additional experiments where we run $\DPI$ with the same state space as $\QWI$, so that the performance differences can be purely attributed to the algorithm rather than the state space. 
}

\subsubsection{Patient-level verification rates.} 
\begin{figure}
\centering
\begin{subfigure}{.47\textwidth}
  \centering
  \includegraphics[width=1\linewidth]{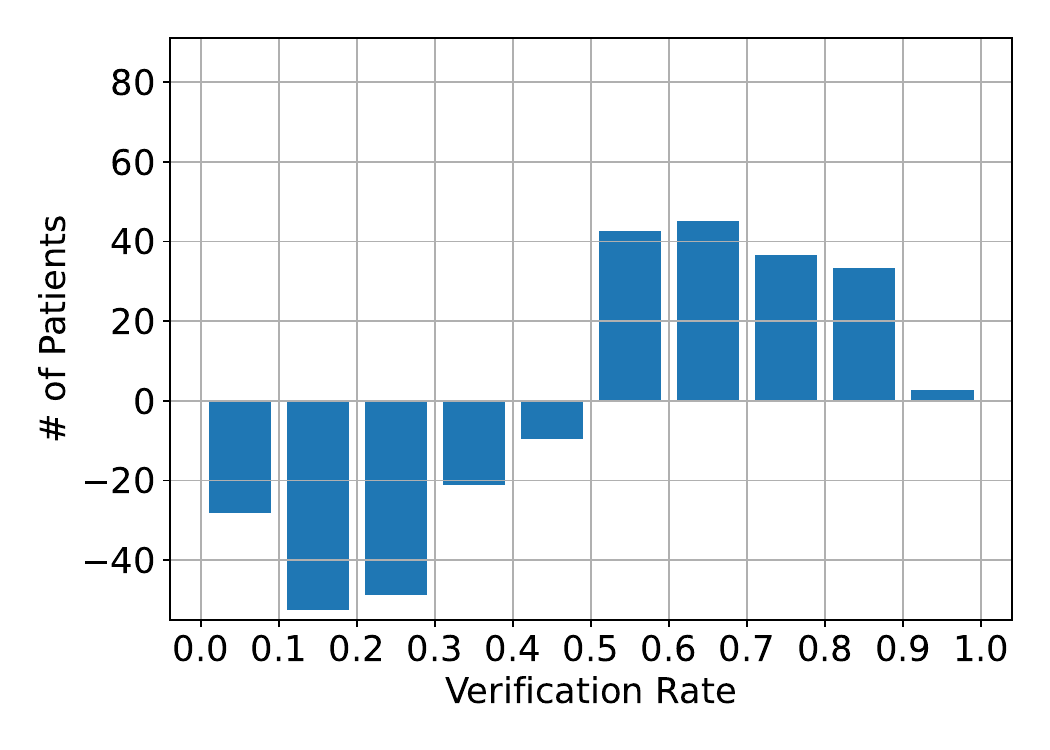}
  \caption{Comparing $\DPI$ to $\baseline$.}
\end{subfigure}%
\begin{subfigure}{.47\textwidth}
  \centering
  \includegraphics[width=1\linewidth]{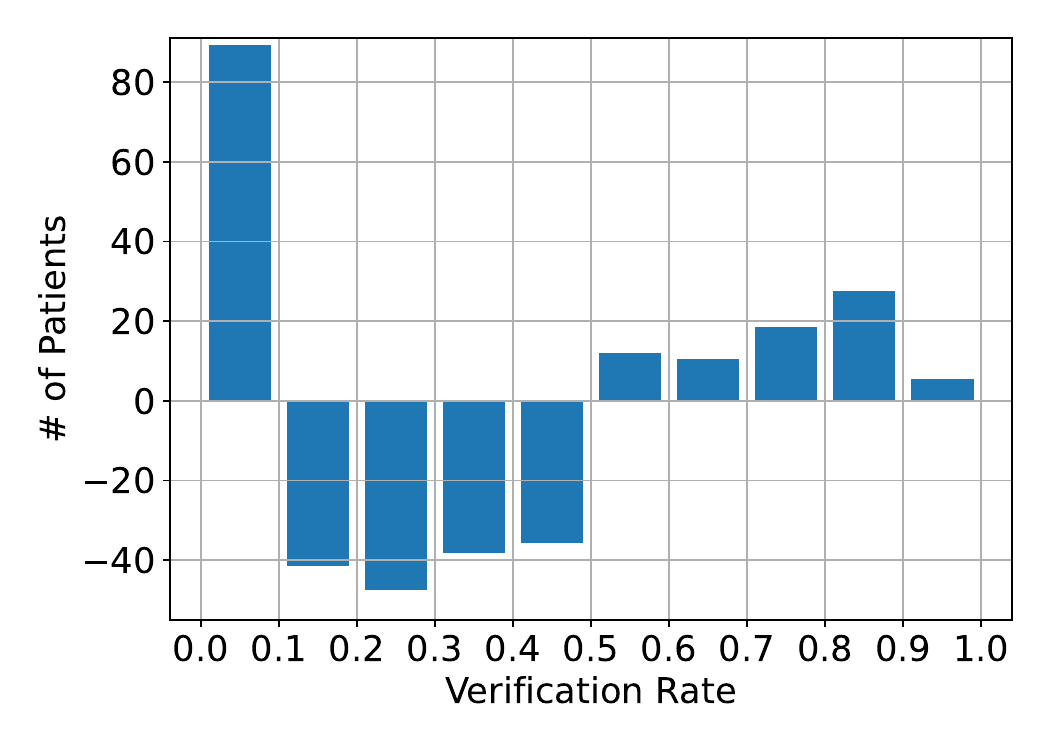}
  \caption{Comparing $\bandit$ to $\baseline$.}
\end{subfigure} \\
\begin{subfigure}{.47\textwidth}
  \centering
  \includegraphics[width=1\linewidth]{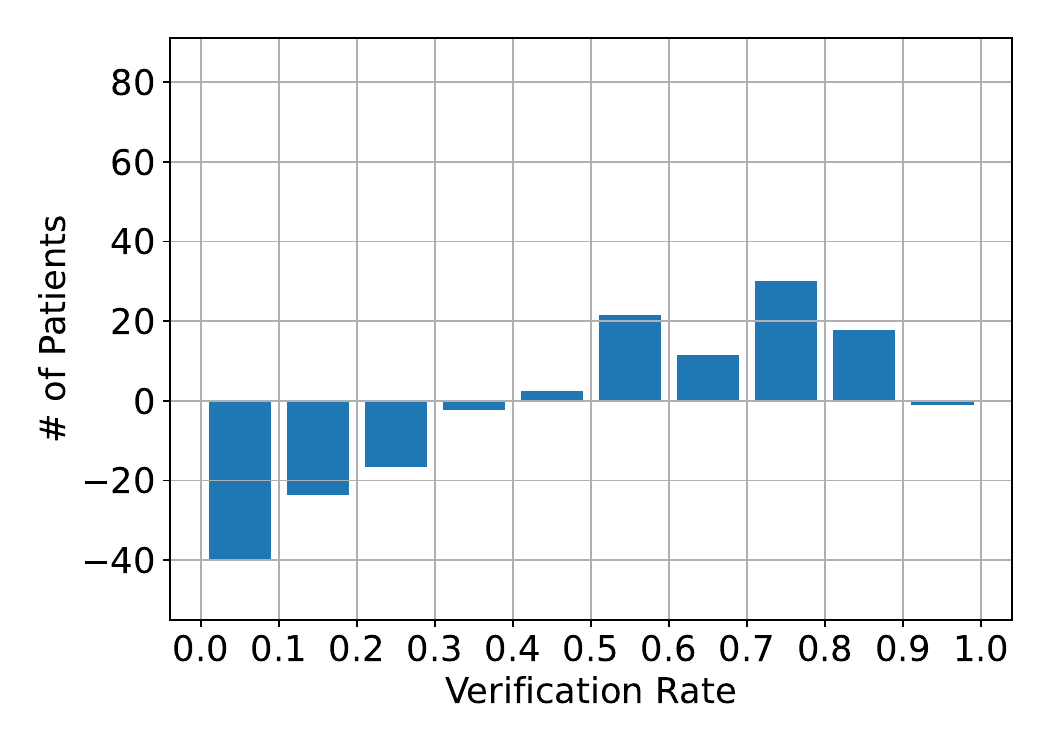}
  \caption{Comparing $\QWI$ to $\baseline$.}
\end{subfigure}
\caption{
\added{
Differences in the distribution of patient verification rates compared to $\baseline$. 
  The bins represent the difference in the number of patients whose overall verification rate is between $0-10\%$, $10-20\%, \dots, 90-100\%$.
For example, the first bin in (a) shows that there were 28 fewer patients whose verification rate was between 0 and 10\% under $\DPI$, compared to $\baseline$.
 There were 3594 patients in total, and the budget was fixed at 26.
 }
}
  \label{fig:diff_vrates_all}
\end{figure}

The overall number of verifications increases under $\DPI$, but how do these rates get impacted at the patient-level?
Fixing the budget to be 26, we compute the verification rate of \textit{each} patient, and we examine the distribution of these patient-level rates. \added{
In Figure~\ref{fig:diff_vrates_all}, we plot how the distribution of patient verification rates shift under the $\DPI$, $\bandit$, and $\QWI$ algorithms, compared to $\baseline$.
We see that under $\DPI$, the distribution shifts in a way that there are fewer patients with verification rates under 50\%, and more patients with a verification rate higher than 50\%.
We see a similar phenomenon for $\QWI$, but the magnitude of the shift is smaller.
$\bandit$ also observes an increase in $>50\%$ verification rates, but there is also an increase of those with very low (0-10\%) verification rates.
These results for $\DPI$ represent a desirable type of shift, where main improvement of $\DPI$ comes from an increase in the number of patients with a high verification rate.
We also provide absolute numbers in \cref{tab.verification_rates}, where we show the percentage of patients whose verification rate is higher than 50\% and 70\% for each of the four algorithms.
Under $\DPI$, the number of patients whose verification rate is above 50\% and 70\% increased relatively by 6.7\% and 5.2\% respectively compared to $\baseline$.
}

\begin{table}[h]
\TableSpaced %
\caption{\added{
The average percentage of patients whose verification rate was over 50\% and 70\% across the four algorithms.
 There were 3594 patients in total, and the budget was fixed at 26.
}} \label{tab.verification_rates}
\vspace{2mm}
\begin{center}
\begin{tabular}{@{}c|cccc@{}}
\toprule
\% patients with \\ verification rate      & \quad $\baseline$ \;  & $\DPI$ & $\bandit$& $\QWI$ \\ \midrule
$\ge 50\%$  &  61.7\%   & 66.1\%  & 63.8\% & 63.9\% \\ 
$\ge 70\%$  &  38.2\%  & 40.2\%  & 39.7\%  & 39.5\% \\ \bottomrule
\end{tabular}
\end{center}
\end{table}

\subsubsection{Description of the targeted patients.} \label{sss.targetedpatients}
In \cref{tab.stats}, we fix the budget to be 26 and we show statistics regarding the state of the targeted patients for each of the four policies.
For example, under $\baseline$, on average, the patient that received an intervention had a treatment effect of 8.8\% with respect to the probability that they will verify the next day.
8.8\% is the `true' average treatment effect, in the sense that the numbers that are averaged are taken directly from the simulation model.

\begin{table}[h]
\TableSpaced %
\caption{
Average statistics of the state of patients who were given an intervention, across the three policies that were run for $B=26$.
(a) is the average value of $f(x, 1) - f(x, 0)$, the increase in probability that the patient verifies the next day when they are given an intervention. 
(b) is the average $f(x, 0)$, the probability that a patient verifies the next day \textit{without} an intervention. 
(c) is the average number of remaining days the patient will be on TB treatment for.
} \label{tab.stats}
\vspace{2mm}
\begin{center}
\begin{tabular}{@{}ccccc@{}}
\toprule
                                                 & $\baseline$ & $\DPI$ & $\bandit$ & $\QWI$ \\ \midrule
\multicolumn{1}{l}{(a) \added{$f(x, 1) - f(x, 0)$}}  & 8.8\%   & 10.6\%    & 13.2\%    & 6.9\%    \\ 
\multicolumn{1}{l}{(b) \added{$f(x, 0)$}}   & 18.2\%  & 22.2\%  & 35.2\%   & 12.2\%      \\ 
\multicolumn{1}{l}{(c) Days on TB treatment remaining} & 68.3     & 109.3   & 92.4    & 69.7      \\ \bottomrule
\end{tabular}
\end{center}
\end{table}

Statistic (a) represents exactly what the $\bandit$ policy optimizes for, the increase in probability of the patient verifying the next day.
The fact that $\bandit$ yields the highest value confirms that indeed, the policy correctly learns what it is supposed to learn.
$\DPI$ chooses patients with a higher one-step treatment effect than $\baseline$, but lower than that of $\bandit$.
Then, the fact $\DPI$ outperforms $\bandit$ in terms of overall verification implies that a myopic strategy of looking only one step ahead is not sufficient.
The next two statistics shed light on why this may be.

\begin{figure}[h]
\begin{center}
\vspace{-5mm}
  \includegraphics[width=0.54\linewidth]{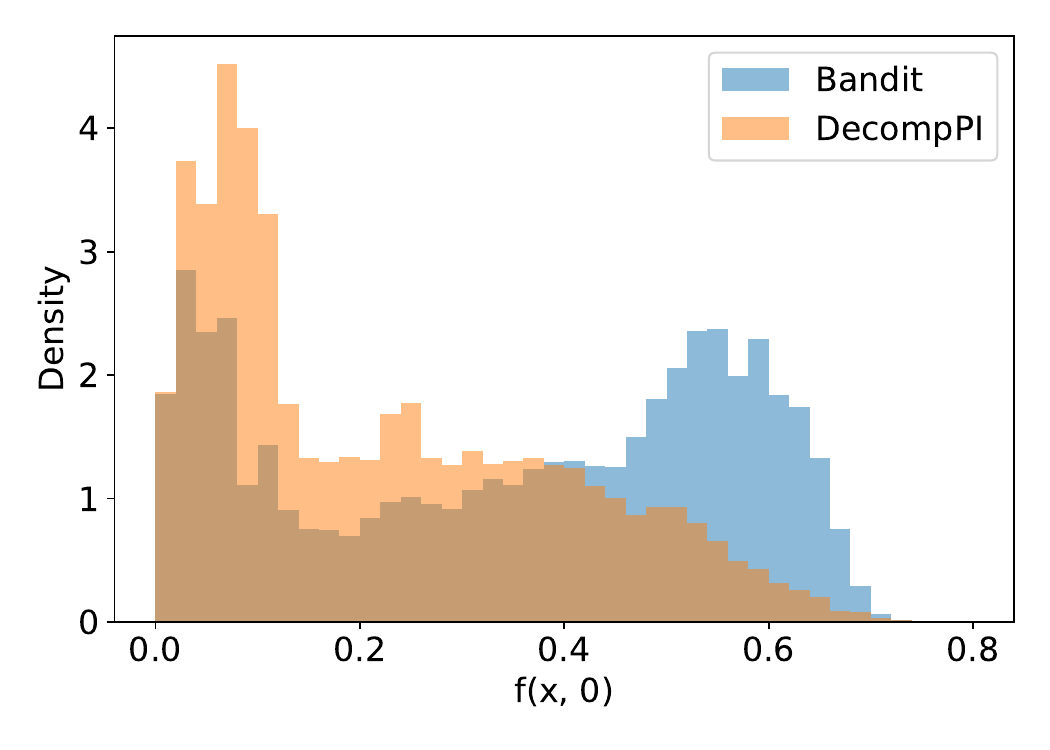}
\vspace{-2mm}
  \caption{
  Histogram of the value of $f(x, 0)$ of targeted patients, the probability that the patients would verify without an intervention.
  This is the entire distribution of the statistic (b) in \cref{tab.stats} for $\bandit$ and $\DPI$.}
  \label{fig:base_probs}
\vspace{-6mm}
\end{center}
\end{figure}

Statistic (b) represents the probability that the targeted patient would have verified anyway without an intervention, and we see that the $\bandit$ targets patients with a much higher verify probability than the other policies. 
We plot the entire distribution of this quantity in Figure~\ref{fig:base_probs}, where we see that $\bandit$ often targets those with a relatively high probability ($>45\%$), while $\DPI$ targets those with a relatively low probability ($<15\%$).
This may contribute to the improved performance of $\DPI$, and the reasoning for this can be seen through the two-state model from \cref{ss.2statemodel}, where statistic (b) corresponds to the parameter $p$.
If two patients have the same values of the parameters $g$ and $\tau$ but differing values for $p$, the intervention value is higher when $p$ is smaller (see \cref{prop:z_clean_form}).
This is because the patient with a high value of $p$ is more likely to switch to state 1 at the current time step as well as all future time steps.
As an extreme example, for a patient with $p=0$, they \textit{need} an intervention to switch to state 1, whereas a patient with $p>0$ may switch to state 1 (either now or in the future), without an intervention.
Therefore, an intervention is more likely to be helpful for those with a smaller value of $p$, which $\DPI$ targets.

On the other hand, $\QWI$ takes the above strategy to an extreme, where it targets those with a very low probability of verifying (12.2\%), but on average these patients also do not have a high next-day treatment effect (6.9\%).
This may explain the inconsistent behavior of $\QWI$ as the budget increases. 
The strategy of targeting these patients with a low verify probability and a low treatment effect is reasonably effective in the mid-budget regime; however, as the budget increases, one may also need to judiciously target other types of patients, which $\QWI$ does not do.

Lastly, statistic (c) is the average number of days a targeted patient has remaining on the platform.
If an intervention positively affects patients for all of their future time steps, then targeting those with longer time left in the system would result in higher benefits. 
The results show that $\DPI$ targets those with the longest days of treatment left.

\subsubsection{Prominent features for $\DPI$.}
\cref{tab.coefs} displays the five most predictive features of the intervention values that $\DPI$ uses to target patients.
These features were found by using Lasso regression with a tuned parameter -- see Section~\ref{s.app.coefficients} for details on the method used.
The results show that the intervention value is lower when the number of previous interventions is higher (first two features), which is intuitive since patients may become fatigued and less receptive when there are too many interventions.
The intervention value is lower when the patient's past verifications is higher (third and fourth features). This is consistent with the analysis in \cref{tab.stats}, where $\DPI$ targets those with a smaller value of $f(x, 0)$.
Lastly, the intervention value increases for patients who are older.

\begin{table}[h]
\TableSpaced %
\caption{
The most predictive features of higher intervention values for $\DPI$, as well as the sign of their coefficient. 
} \label{tab.coefs}
\vspace{2mm}
\begin{center}
\begin{tabular}{@{}lc@{}}
\toprule
\; Feature & Sign of Coefficient  \\ \midrule
\; Interventions: total number & $-$ \\ 
\; Interventions: \# previous week &  $-$ \\
\; Verifications: overall percentage &  $-$ \\
\; Verifications: \# previous week & $-$ \\
\; Age & $+$ \\
\bottomrule
\end{tabular}
\end{center}
\end{table}

\subsection{\added{Robustness of State Space Representation}} \label{ss.state_space_results}
\added{One of the factors attributing to the performance gap between $\DPI$ and $\QWI$ is the differences in state space representation. $\QWI$ requires a finite state space and its computation time scales with the number of states. Hence, we use a relatively small state space for $\QWI$ for our numerical experiments. The ease of using a larger and infinite size state space is an inherent advantage of $\DPI$ over $\QWI$; however, in order to isolate the performance difference caused by the algorithm itself, we run $\DPI$ using the same state space as $\QWI$.

We try two variants of $\DPI$ that differ in the state space used:
\begin{itemize}
	\item \textsf{DecompPI-3}: This uses the same three features used for $\QWI$ (number of times verified in the last week, total historical verification rate, and number of interventions received in the last week), but these features are \textit{not discretized}, and hence the size of the state space is still infinite.
	\item \textsf{DecompPI-3-discrete}: This uses the exact same state space as $\QWI$ --- three features that are discretized in the same way, resulting in 45 states.
\end{itemize}

\begin{figure}[h]
\begin{center}
\vspace{-3mm}
  \includegraphics[width=1\linewidth]{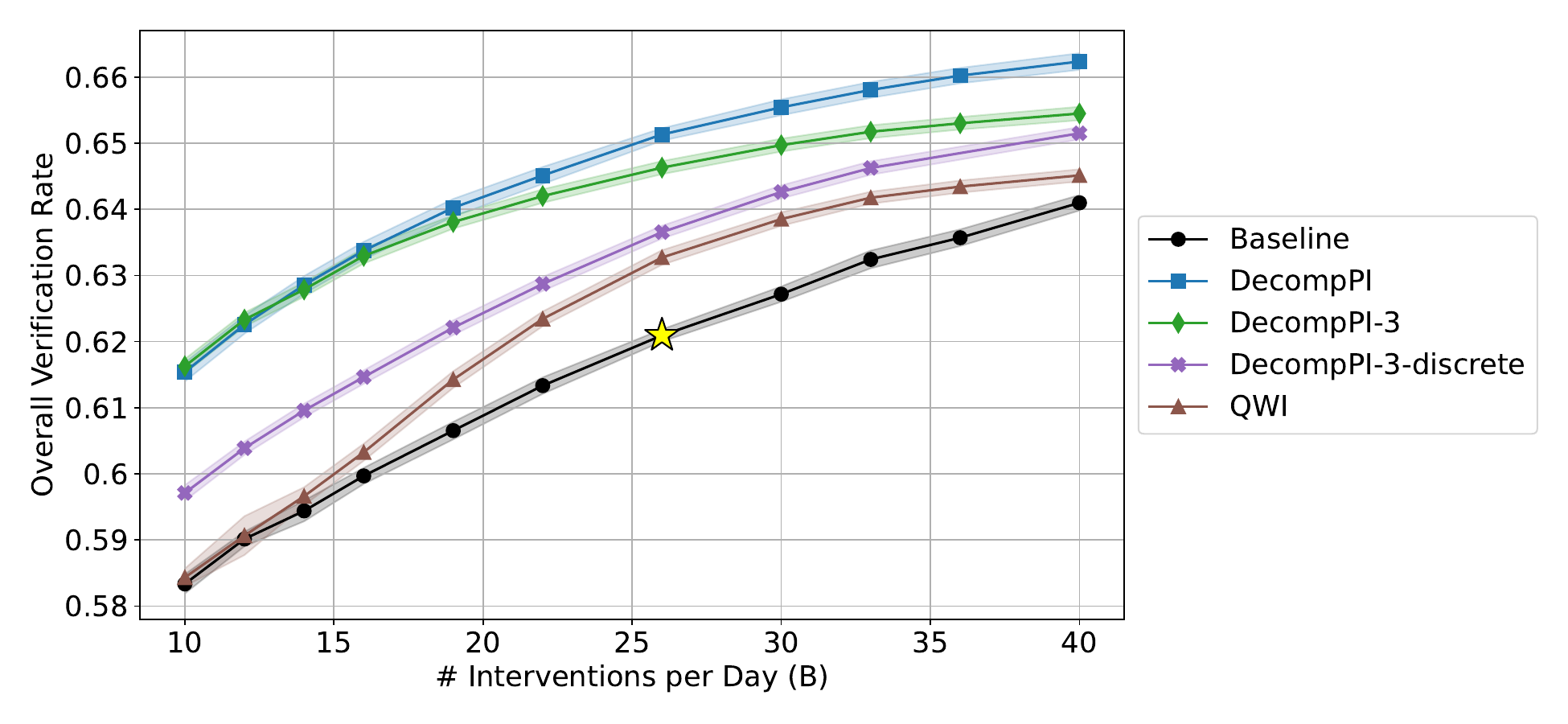}
  \caption{Average overall verification rate over 50 runs for each policy and budget. The shaded region indicates a 95\% confidence interval. The star represents the operating point for Keheala.}
  \label{fig:DPI3}
\vspace{-6mm}
\end{center}
\end{figure}

The performance of these two algorithms, along with the original $\DPI$ and $\QWI$ policies are shown in Figure~\ref{fig:DPI3}.
The policies \textsf{DecompPI-3-discrete} and $\QWI$ use the exact same state space, and we see that the former consistently outperforms the latter.
This comparison isolates the performance gap induced by the \emph{algorithms}, and the results provide robust evidence on the strength of $\DPI$.
Lastly, we see that \textsf{DecompPI-3} consistently has a strong performance, comparable to that of $\DPI$.
This demonstrates the robustness of $\DPI$ with respect to the feature space, and it also exemplifies the benefit of employing an infinite state space compared to a discretized one.  
}

\section{Conclusions, Implications, and Future Directions}\label{s.conclusion}

This work tackles an important problem of personalizing and optimizing costly interventions in the context of digital systems for behavioral health.  
We develop an approach, $\DPI$, that learns an intervention policy from an existing dataset collected from a pilot study.
 $\DPI$ is model-free, which avoids the need to specify a model of patient behaviors.
Unlike many reinforcement learning methods which often rely on a long horizon to achieve good performance, $\DPI$ leverages offline data to immediately provide an effective policy.
We provide a theoretical guarantee on a special case of the model that is a stylized representation of the practical setting of interest, and it exhibits strong empirical performance on a validated simulation model of a real-world behavioral health setting.

\subsection{Managerial Implications}
\added{The operational settings that could benefit most from our work are those resembling our motivating example of \textit{Keheala}. In other words, consider an organization that has designed a protocol for a new behavioral program, by which some subjects are eligible for a costly intervention. A pilot implementation of this program reveals severe resource limitations such that it is infeasible to provide the intervention to all eligible subjects. During this pilot, the organization provides the intervention in an ad hoc manner (among eligible subjects) and collects data on engagement and outcomes. Our analysis reveals that as long as the ad hoc provision of the intervention during the pilot phase is practically random (e.g., if the intervention is delivered based on a call list that is alphabetical), $\DPI$ can be used to improve performance in subsequent deployment or scale-up, through additional targeting within the set of eligible subjects. Furthermore, our theoretical and numerical analysis consistently demonstrate that the value of $\DPI$ increases as the available resources become more limited, which is important since (as we discuss in \cref{s.intro}) algorithms that rely on online experimentation can be inefficient or even counterproductive in such settings. 

The above description clearly does not apply to all behavioral health settings, but we believe it is relevant for many new initiatives that are being designed, piloted, and scaled-up. In particular, we believe $\DPI$ can be valuable for interventions that aim to serve disadvantaged groups or regions since those inherently face more stringent resource limitations.}

\subsection{Limitations and Future Directions}
Lastly, we discuss limitations of the current work that serve as valuable future directions.
One gap between our model and the practical application of Keheala is that the ultimate objective of Keheala is to improve eventual health outcomes (i.e., cure patients of TB).
There are two major hurdles that need to be addressed in order to fully align with this goal, within the existing infrastructure of Keheala (of using daily adherence information).
The first obstacle is the lack of a mapping between treatment adherence patterns to health outcomes.
It has been shown that higher verification rates are associated with better outcomes \citep{Boutilier22Improving}
but it would be valuable to identify more specific and causal relationships (e.g., is it more important for a patient to adhere to their treatment in the earlier phase of their treatment regime compared to later?).
Addressing this issue is very specific to TB but would be tremendously valuable not only for optimizing a platform like Keheala, but also for the broader medical research on TB.
Second, a problem less specific to TB, is to design policies that can optimize for reward functions which are not necessarily additive for each time step (e.g., maximize the number of patients whose overall verification percentage is over 70\%).

Next, there are other interesting extensions to the model and the algorithm that one can consider.
With respect to performance guarantees on the stylized model in \cref{s.theory}, it would be valuable to analyze how the guarantee is impacted by various modeling extensions such as generalizing the MDP (e.g., more states, having interventions impact all states) or extending the class of baseline policies. 
From an algorithmic standpoint, valuable extensions include incorporating online samples, and incorporating other practical considerations such as fairness in how the interventions are distributed across patients. \added{Finally, even if the time horizon for TB treatment is determined by the drug regimen length, one can imaging other behavioral health situations in which the time horizon used to calculate the intervention value for each subject is endogenous to their prior behavior, such that increased engagement results in reduced likelihood of leaving the service.}

\ACKNOWLEDGMENT{The authors are grateful to Jon Rathauser, founder and CEO of Keheala, for the collaboration. The authors also thank the Keheala Operational Team (Alice, Faith, Edwin, Jacinta, Jill, Lewis, Moreen, Trish).}

\vspace*{-0.5em}
\bibliographystyle{ormsv080}
\bibliography{Keheala}

\clearpage

\setcounter{page}{1} \renewcommand{\theequation}{A\arabic{equation}}
\setcounter{equation}{0}
\renewcommand{\thelemma}{A\arabic{lemma}}
\setcounter{lemma}{0}
\renewcommand{\theproposition}{A\arabic{proposition}}
\setcounter{proposition}{0}
\renewcommand{\thesection}{A\arabic{section}}
\setcounter{section}{0}
\renewcommand{\thefigure}{A\arabic{figure}}
\setcounter{figure}{0}
\renewcommand{\theremark}{A\arabic{remark}}
\setcounter{remark}{0}
\renewcommand{\thetable}{A\arabic{table}}
\setcounter{table}{0}

\section*{E-companion for ``Policy Optimization for Personalized Interventions in Behavioral Health''}

~

\section{Table of Notation} \label{s.app.tablenotation}

\begin{table}[h]
\TableSpaced %
\caption{
\added{List of notation used in the paper and the proofs in \cref{s.app.proof}.}
} \label{tab.notation}
\vspace{2mm}
\begin{center}
\begin{tabular}{@{}cl@{}}
\toprule
Notation & Meaning\\ \midrule
$N$ & Number of patients \\ 
$T$ & Number of time steps \\ 
$B$ & Budget \\ 
$\cS$ & State space for the MDP of a single patient\\ 
$\cA$ & Action space for the MDP of a  single patient\\ 
$\cA^N$ & Action space for the system MDP \\ 
$\mathcal{M}_i$ & MDP for patient $i$ \\
$\mathcal{P}_i(S, S', A)$ & Transition probability function for patient $i$ \\
$R_i(S, S', A)$ & Reward function for patient $i$ \\
$S_{it}^\pi$ & State of patient $i$  at time $t$ under policy $\pi$ \\
$A_{it}^\pi$ & Action for patient $i$  at time $t$ under policy $\pi$ \\
$q_{it}^{\pi}(S, A)$ & Patient-level $q$-value defined in \eqref{eq:q} \\
$\bS, \bA$ & Vector of states and actions for the system-level MDP \\
$Q_{t}^\pi(\bS, \bA) $ & System-level $Q$-value defined in \eqref{eq:Q} \\
$z_{it}^{\pi}(S)$ & Intervention value used for $\DPI(\pi)$ \\
$p_i, g_i, \tau_i$ & Parameters for the two-state model of \cref{ss.2statemodel} \\
$\gamma$ & Probability of an intervention for the policy $\RAND(\gamma)$ \\
$\bar{M}$ & Instance-dependent parameter used in \cref{thm:approx_ratio_b} \\
$V_{it}$ & Whether patient $i$ verifies at time $t$ \\
$P_{it}, G_{it}, K_{it}, W_{it}$ & Bernoulli variables used in the sample path coupling of \cref{sec:coupling} \\
$\bar{z}_{it}(S)$ & Null intervention values defined as $\lim_{\gamma \to 0^+} z_{it}^{\gamma}(S)$ \\
$\DPI_0$ & Index policy that ranks patients by the null intervention values \\
$I_{t}$ & Subset of patients in state 0 at time $t$ \\
$D_t(I)$ & Subset of patients that $\DPI_0$ would choose out of $I \subseteq [N]$ \\
$\ONE(i, t)$ & Policy that chooses patient $i$ once at time $t$ \\
$\tilde{S}_{ir|t}$ & State of patient $i$ at time $r$ under $\ONE(i, t)$ \\
$Z_{it}$ & Number of times state of patient $i$ differ in $\ONE(i, t)$ compared to $\NULL$, defined in \eqref{eq:define_Y} \\
$\nu_i(t)$ & Last time before $t$ that $i$ was in state 0, defined in \eqref{eq:definetau} \\
\bottomrule
\end{tabular}
\end{center}
\end{table}

\section{Proof of \cref{thm:approx_ratio_b} and \cref{corr:approx}}  \label{s.app.proof}

We first prove \cref{thm:approx_ratio_b}, where we break down the proof into four steps, where the bulk of the proof lies in the first two steps.
The proof of \cref{corr:approx} follows from steps 3 and 4, which we describe in \cref{sec:app:pf_approx}.
The proofs of all of the steps make use a specific sample path coupling procedure which we describe in the next subsection.

\paragraph{Step 1.}
First, we define \textit{null intervention values} as $\nz_{it}(s) = \lim_{\gamma \to 0^+} z^{\gamma}_{it}(s)$. 
We show that for any algorithm $\ALG$, the difference $\ALG - \NULL$ can be written as a sum of the null intervention values of the patients that were chosen.
\begin{proposition} \label{prop:alg_minus_base}
For any algorithm $\ALG$,
\begin{align} \label{eq:diff_char}
\ALG - \NULL = \bE\left[\sumT \sum_{i \in A^{\ALG}_t} \nz_{it}(0) \right].
\end{align}
\end{proposition}

\paragraph{Step 2.}
Then, we define the policy $\DPI_0$ to be the policy which orders patients in state 0 with respect to the null intervention values, $\nz_{it}(0)$, gives interventions to the $B$ patients with the highest values.
We show that this policy achieves at least half of the optimal improvement over $\NULL$.
\begin{proposition} \label{prop:dpinull_half_result}
For any instance of the two-state model,
\begin{align} \label{eq:approx_ratio_b}
\DPI_0  - \NULL \;\geq\; \frac{1}{2} \; ( \OPT - \NULL ).
\end{align}	
\end{proposition}

\paragraph{Step 3.}
Next, we show that if a policy uses an index rule with indices that approximate the null intervention values, this policy also yields a performance guarantee. 
\begin{proposition} \label{prop:approx_index}
Fix $\alpha_1 \in (0, 1]$ and $\alpha_2 \geq 1$.
Let $\ALG$ be an index policy that uses indices $z^{\ALG}_{it}(0)$ that satisfy $\alpha_1 \nz_{it}(0) \leq z^{\ALG}_{it}(0) \leq \alpha_2 \nz_{it}(0)$ for all $i$ and $t$. Then,
\begin{align} \label{eq:approx_ratio_b}
\ALG  - \NULL \;\geq\; \frac{\alpha_1}{2\alpha_2} \; ( \OPT - \NULL ).
\end{align}	
\end{proposition}

\paragraph{Step 4.}
Lastly, we show that the intervention values $z^{\gamma}_{it}(0)$ satisfy the above, where the corresponding $\alpha$ is a function of both $\gamma$ and the underlying parameters.
\begin{proposition} \label{lemma:rand_null}
Fix a patient $i$, and let $M_i = \frac{\tau_i (1-p_i-g_i)}{(p_i + g_i)(1-p_i)} > 0$.
For any $t \in [T]$, the intervention values $z^{\gamma}_{it}(0)$ and $\nz_{it}(0)$ satisfy the following relationship:
\begin{align}
\frac{1}{1 + \gamma M_i} \nz_{it}(0) \; \leq \; z^{\gamma}_{it}(0) \; \leq \; \nz_{it}(0).
\end{align}
\end{proposition}

Using \cref{lemma:rand_null}, we can apply \cref{prop:approx_index} using $\alpha_1 = 1/(1+\gamma M)$ and $\alpha_2 = 1$ for $M = \max M_i$, which completes the proof of \cref{thm:approx_ratio_b}.
The proofs of Propositions \ref{prop:alg_minus_base}-\ref{lemma:rand_null} can be found in Sections \ref{s.app.p1}-\ref{s.app.p4}.

\paragraph{Terms and notation.} We define a couple of terms and notation that we will use for the proofs.
We say that a patient is `chosen' to mean that the patient received an intervention, and we say that a patient is `available' to mean they are in state 0.
Let $I_t \subseteq [N]$ be the set of patients that are in state  0 at time $t$. Then, $[N] \setminus I_t$ are the patients in state 1 at time $t$.
We use $A_t \subseteq [N]$ to refer to the subset of patients who are chosen (instead of using the notation $A_{it} \in \{0, 1\}$ to denote whether patient $i$ is chosen).
Without loss of generality, we assume that $A_t \subseteq I_t$. 
This is because giving an intervention to a patient in state 1 has no impact on their transitions.
We put $\pi$ in the superscript, $I^\pi_t$ and $A^\pi_t$, to refer to the random variables induced by running policy $\pi$.

We now describe the sample path coupling, and then prove each of the above four propositions in the following subsections.

\subsection{Sample path coupling} \label{sec:coupling}
Fix $\gamma \in (0, 1]$.
We specify a new set of model dynamics that couples different policies through shared random variables.
We will show that these new dynamics are equivalent to the original dynamics specified in \cref{ss.2statemodel}.

\noindent
\textbf{New dynamics.}
At each time $t=1, \dots, T-1$, the following occurs:
\begin{enumerate}
	\item States $S_{it}$ are observed for all patients $i$.
	\item A policy selects a subset of patients $A_{t} \subseteq I_t$ for an intervention.
	\begin{itemize}
		\item For the policy $\RAND(\gamma)$, draw $W_{it} \sim \Bern(\gamma)$ independently for each patient $i$.
		Then, $i \in A_t$  if and only if $S_{it} = 0$ and $W_{it} = 1$.
	\end{itemize}
	\item For every patient $i \in [N]$, draw independent, Bernoulli variables
	$G_{it} \sim \Bern(g_i / (1-p_i))$, $P_{it} \sim \Bern(p_i)$, $K_{it} \sim \Bern(\tau_i/(1-p_i))$.
	\item Patient transitions occur in the following order:
\begin{enumerate}[label=(\roman*)]
	\item \textbf{Return to state 0:} For each patient $i \in I_t(1)$ and $G_{it} = 1$, the patient returns to state 0.
	Let $I_t' = I_t \cup \{i : G_{it} = 1, i \in I_t(1)\}$ be the new set of patients in state 0.
	\item \textbf{Passive transitions to state 1:}  For each patient $i \in I_t'$ and $P_{it} = 1$, the patient goes to state 1.
	Let $I_t'' = I_t' \setminus \{i : P_{it} = 1, i \in I_t'\}$ be the new set of patients in state 0.
	\item \textbf{Active transitions to state 1:}  For each patient $i \in I_t'' \cap A_{t}$ with $K_{it} = 1$, the patient goes to state 1.
	Then, $I_{t+1} = I_t'' \setminus \{i : K_{it} = 1, i \in I_t'' \cap A_{t}\}$ are the set of patients in state 0 at the next time step, and $I_{t+1}(1) = [N] \setminus I_{t+1}$.
\end{enumerate}
	\item We collect the reward $I_{t+1}(1)$.
\end{enumerate}

\textbf{Equivalence.}
We claim that the original model dynamics (from \cref{ss.2statemodel}) is equivalent to these new model dynamics.
To show this, we need to show that the transition probabilities between states are equal under both models.

In the new model, suppose a patient is in state 0.
If they were not chosen, they can only transition to 1 under step (ii), which occurs with probability $p_i$.
If they were chosen, they can transition to 1 under either step (ii) or (iii).
In total, they transition with probability $p_i + (1-p_i) \cdot \tau_i/(1-p_i) = p_i + \tau_i$.
Next, suppose a patient is in state 1.
For the patient to transition to state 0, it must be that $G_{it} = 1$ and $P_{it} = 0$.
This occurs with probability $g_i / (1-p_i) \cdot (1-p_i) = g_i$.
Therefore, the transition probabilities between the two models are equal for all patients.

\textbf{Coupling sample paths of policies.} 
From now on, we assume that all policies are coupled through the variables $P_{it}, G_{it}, K_{it}$.
This coupling immediately gives us the following properties.

\begin{lemma} \label{prop:coupling_property0}
For any policy, if $P_{it} = 1$, then $S_{i, t+1} = 1$.
\end{lemma}

\begin{myproof}
Suppose $P_{it} = 1$ for some $i$ and $t$.
In step 4(ii) of the new dynamics, the set $I_t''$ is defined so that $i \in I_t''$.
In step 4(iii), $I_{t+1}$ is a subset of $I_t''$,  and $I_{t+1}(1) = [N] \setminus I_{t+1}$.
Hence $S_{i, t+1} = 1$.
\end{myproof}

The next property says that if a patient is in state 1 under the $\NULL$ policy, they must also be in state 1 under any other policy.

\begin{lemma} \label{prop:coupling_property}
Let $\ALG$ be any policy. If $S_{it}^{\NULL} = 1$, then $S_{it}^{\ALG} = 1$.
Equivalently, if $S_{it}^{\ALG} = 0$, then $S_{it}^{\NULL} = 0$.
\end{lemma}

\begin{myproof} %
Let $i, t$ be such that $S_{it}^{\NULL} = 1$.
Let $t' = \max\{t' < t : P_{it'} = 1\}$ be the most recent time $P_{it'}$ was 1.
Since $S_{it}^{\NULL} = 1$, $G_{is} = 0$ for every $s \in \{t'+1, \dots, t-1\}$.
Since $P_{it'} = 1$, \cref{prop:coupling_property0} implies $S^{\ALG}_{i, t'+1} = 1$.
Since $G_{is} = 0$ for every $s \in \{t'+1, \dots, t-1\}$, $S_{it}^{\ALG} = 1$.
\end{myproof}

Lastly, using this sample path coupling, we can write an expression for the intervention values $z^{\gamma}_{it}(0)$ and $\nz_{it}(0)$.
\begin{lemma} \label{lemma:zgamma}
The intervention value with respect to $\RAND(\gamma)$ can be written as 
\begin{align} \label{eq:zformula}
z^{\gamma}_{it}(0) = \tau_i \cdot \bE[\min\{\text{Geometric}(p_i + g_i + \gamma \tau_i (1-p_i-g_i)/(1-p_i)), T-t+1\}].
\end{align}
Moreover, the null intervention value can be written as
\begin{align} \label{eq:nzformula}
\nz_{it}(0) = \tau_i \cdot \bE[\min\{\text{Geometric}(p_i + g_i), T-t+1\}].
\end{align}
\end{lemma}

\begin{myproof}[Proof of \cref{lemma:zgamma}]
Recall that $z^{\gamma}_{it}(0) = g_{it}^{\RAND(\gamma)}(0, 1) - g_{it}^{\RAND(\gamma)}(0, 0)$.
Let $E_0 = \{S_{it} = 0, A_{it} = 0\}$ be the event where patient $i$ is in state 0 at time $t$ and $\RAND(\gamma)$ does not choose the patient and let $E_1 = \{S_{it} = 0, A_{it} = 1\}$ be the event where patient $i$ is in state 0 at time $t$ and $\RAND(\gamma)$ does choose the patient.

Conditioned on either $E_0$ or $E_1$, the distribution of the variables $W_{it'}$ for $t' > t$ and  $P_{it'}, G_{it'}, K_{it'}$ for $t' \geq t$ are the same, due to independence.
Therefore, to compute $z^{\gamma}_{it}(0)$, we consider two hypothetical sample paths for patient $i$, one where $E_0$ holds and one where $E_1$ holds, but where all future variables are exactly the same.
We refer to $S^0_{it'},S^1_{it'} \in \{0, 1\}$ as the states under the two sample paths respectively at time $t' > t$.
Let $\zeta = \min\{T+1, \min\{t' > t: S^0_{it'} = S^1_{it'}\}\}$ be the first time after time $t$ that the two states converge, where $\zeta = T+1$ if they never converge. 
The states will always be the same after time $\zeta$ due to the sample path coupling.
Then, $z^{\gamma}_{it}(0) = \bE[\zeta] - t - 1$.

We now write an expression for $z^{\gamma}_{it}(0)$.
For the states to differ at time $t+1$, it must be that $P_{it} = 0$ and $K_{it} = 1$.
After that, the states will converge at time $t'+1$ if (i) $P_{it'} = 1$,  (ii) $G_{it'} = 1$, or (iii) $W_{it'} = 1$ and $K_{it'} = 1$.
Let $\Gamma = \min\{t' > t : P_{it'} = 1 \} - t$ be the length of time from $t$ until $P_{it'} = 1$.
Let $\Gamma' = \min\{t' > t : G_{it'} = 1 \} - t$ be length of time from $t$ until $G_{it'} = 1$.
Let $\Gamma'' = \min\{t' > t : W_{it'} = 1, K_{it'} = 1 \} - t$ be length of time from $t$ until $W_{it'} = 1$ and $K_{it'} = 1$.
Then, $z^{\gamma}_{it}(0)$ can be written as
\begin{align*}
z^{\gamma}_{it}(0) = \Pr(K_{it} = 1) \cdot \Pr(P_{it} = 1) \cdot \bE[\min\{\Gamma, \Gamma', \Gamma'', T-t+1\}].
\end{align*}
The term $\min\{\Gamma, \Gamma', \Gamma''\}$ is a geometric random variable with parameter 
\begin{align*}
&1-(1-\Pr(P_{it} = 1))(1-\Pr(G_{it} = 1))(1-\Pr(W_{it} = 1, K_{it} = 1)) \\
=& 1-(1-p)(1-g/(1-p))(1-\gamma \tau / (1-p)) \\
=& p + g + \gamma \tau (1-p-g)/(1-p).
\end{align*}
Therefore,
\begin{align*} 
z^{\gamma}_{it}(0) = \tau \cdot \bE[\min\{\text{Geometric}(p + g + \gamma \tau (1-p-g)/(1-p)), T-t+1\}].
\end{align*}
\cref{eq:nzformula} follows from taking the limit of the above as $\gamma \to 0$, using the dominated convergence theorem.
\end{myproof}

\subsection{Step 1: Proof of \cref{prop:alg_minus_base}} \label{s.app.p1}
We start with analyzing the left-hand side, $\ALG - \NULL$.
\cref{prop:coupling_property} says that whenever $S^{\ALG}_{it} = 0$, it must be that $S^{\NULL}_{it} = 0$.
Therefore, $\ALG-\NULL$ can be written as the number of times when $S^{\ALG}_{it} = 1$ while $S^{\NULL}_{it} = 0$:
\begin{align}
\ALG - \NULL = \bE\bigg[\sumT \bI(S^{\ALG}_{it'} =1, S^{\NULL}_{it'} = 0) \bigg]
\end{align}

Due to the sample path coupling, $S_{it}^{\ALG} \neq S_{it}^{\NULL}$ can only occur if
$\ALG$ chose patient $i$ at a prior time step, and the states have been different since then (if the states converged, it will stay the same unless $\ALG$ chose the patient again).
Therefore, each time $\bI(S^{\ALG}_{it'} =1, S^{\NULL}_{it'} = 0)$ occurs, it is associated with an intervention by $\ALG$ at a previous time step.
Hence, we will instead represent $\ALG-\NULL$ by summing over all interventions given by $\ALG$, and relating each intervention to how long the states $S^{\ALG}_{it'}$ and $S^{\NULL}_{it'}$ deviate.

\textbf{Defining counterfactual state.}
To formalize this notion, we need to define a couple of new quantities that will play a important role in both this step and step 2 of the proof.
We define a policy $\ONE(i, t)$ to be the same as the $\NULL$ policy, except that it chooses patient $i$ once at time $t$.
Define $\tilde{S}_{i r | t} = S^{\ONE(i, t)}_{ir}$ to be the state of patient $i$ at time $r$ under this policy, which we call the \textit{counterfactual state}.
Then, let $Z_{it}$ be the number of times that the counterfactual state is not equal to the state under $\NULL$:
\begin{align} \label{eq:define_Y}
Z_{it} &= |\{t' \in [T] \; : \; S_{it'}^{\NULL} \neq \tilde{S}_{it' | t} \}|.
\end{align}

Note the following properties:
\begin{itemize}
	\item The two states are always equal before time $t$.
	($S_{it'}^{\NULL} =\tilde{S}_{it' | t}$ for any $t' \leq t$.)
	\item Once the two states converge at some time $t' > t$, they will never diverge again since the policies are the same. (If $S_{it'}^{\NULL} =\tilde{S}_{it' | t}$ for some $t' > t$, then the same holds for any $r > t'$.)
	\item The only way that the states can be different is if the state is 0 under $\NULL$ and 1 under $\ONE(i, t)$, due to \cref{prop:coupling_property}.
\end{itemize}
Therefore, $Z_{it}$ represents exactly the increase in total reward from patient $i$ caused by the intervention at time $t$, compared to $\NULL$.
More specifically, when $S^{\NULL}_{it} = 0$, $Z_{it}$ is the number of time steps that the patient was in state 1 right after time $t$, before it transitioned back to state 0 or the state under the $\NULL$ policy also moved to state 1 (or we reached the last time step).
The above logic allows us to write out an expression for the expected value of $Z_{it}$, conditioned on $S^{\NULL}_{it} = 0$.

\begin{lemma} \label{lemma:zequalsy}
$\bE\big[Z_{it} \;|\; S^{\NULL}_{it} = 0 \big] = \tau_i \cdot \bE[\min\{\text{Geometric}(p_i + g_i, T-t+1\}] = \nz_{it}(0)$.
\end{lemma}

Now, we can write $\ALG-\NULL$ as the sum of $Z_{it}$ values at the times when $i$ was chosen:
\begin{align}
\ALG - \NULL 
&= \bE\bigg[\sumT \sum_{i \in A_t^{\ALG}} Z_{it}\bigg] \label{eq:propRandom} \\
&= \sumT \sum_{i \in [N]} \bE\big[\bI(i \in A_t^{\ALG}) Z_{it}\big] \nonumber \\
&= \sumT \sum_{i \in [N]} \bE\big[\bI(i \in A_t^{\ALG})\big]\; \bE\big[Z_{it} \;|\; i \in A_t^{\ALG}\big]  \nonumber \\
&= \sumT \sum_{i \in [N]} \bE\big[\bI(i \in A_t^{\ALG})\big] \nz_{it}(0) \nonumber \\
&= \bE\bigg[\sumT \sum_{i \in A_t^{\ALG}}  \nz_{it}(0) \bigg], \label{eq:propDet} 
\end{align}
as required.

\begin{myproof}[Proof of \cref{lemma:zequalsy}]
Let $\Gamma_{it} = \min\{t' > t : P_{it'} = 1 \} - t$ be the length of time from $t$ until $P_{it'} = 1$.
Let $\Gamma'_{it} = \min\{t' > t : G_{it'} = 1 \} - t$ be length of time from $t$ until $G_{it'} = 1$.
Then, 
\begin{align} \label{eq:Z_explicit}
Z_{it} = \bI(S_{it}^\NULL = 0, P_{it} = 0, K_{it} = 1) \min\{\Gamma_{it}, \Gamma'_{it}, T-t+1\}.
\end{align}
The indicator represents the fact that $i$ would not transition to state 1 under $\NULL$ ($P_{it} = 0$), but it would transition under $\ONE(i, t)$ ($K_{it} = 1$).
That is, at time $t+1$, the patient is in state 1 under $\ONE$ but in state 0 under $\NULL$.
The $\min\{\Gamma_{it}, \Gamma'_{it}\}$ term counts how long this is the case.
This could end either because patient transitions to state 1 under $\NULL$ (captured by $B_{it}$), or it could be that the patient in $\ONE$ transitions back to state 0 (captured by $L_{it}$).

Note that $\min\{\Gamma_{it}, \Gamma'_{it}\}$ is a geometric random variable with parameter $1-(1-\Pr(P_{it} = 1))(1-\Pr(G_{it} = 1)) = p_i+g_i$.
Therefore,
\begin{align*}
\bE[Z_{it} \;|\; S_{it}^\NULL = 0] 
&= \Pr(P_{it} = 0) \cdot \Pr(K_{it} = 1) \; \bE[\min\{\text{Geometric}(p_i+g_i), T-t+1\}] \\
&= \tau_i \cdot \bE[\min\{\text{Geometric}(p_i+g_i), T-t+1\}].
\end{align*}
Note that the above expression is the same as the one for $\nz_{it}(0)$ from \cref{lemma:zgamma}.
\end{myproof}

\subsection{Step 2: Proof of \cref{prop:dpinull_half_result}}
We prove a more general and stronger version of \cref{prop:dpinull_half_result}, which we state as \cref{thm:step2stronger}.

Define the policy $\DPI_0$ to be the policy which orders patients in state 0 with respect to the null intervention values, $\nz_{it}(0)$, gives interventions to the $B$ patients with the highest values.
Denote by $D_t(I_t^{\ALG}) \subseteq I_t^{\ALG}$ the subset of patients that $\DPI_0$ \textit{would choose} out of $I_t^{\ALG}$, the $B$ patients with the highest null intervention values, $\nz_{it}(0)$.
The next result shows that the sum of null intervention values of the patients in $D_t(I_t^{\ALG})$, will lead to at least half of total sum intervention values for an optimal policy.
\begin{proposition} \label{thm:step2stronger}
For any $\ALG$, 
\begin{align} \label{eq.main_result}
\OPT - \NULL \leq 2 \bE\left[ \sumT \sum_{i \in D_t(I_t^{\ALG})} \yit(0) \right].
\end{align}
\end{proposition}
Note that if $\ALG = \DPI_0$, then $D_t(I_t^{\ALG})$ is simply the patients that $\DPI_0$ chooses at time $t$; in that case, the RHS of \eqref{eq.main_result} equals the RHS of \eqref{eq:diff_char}.
Then, by \cref{prop:alg_minus_base}, the RHS of \eqref{eq.main_result} equals $\DPI_0 - \NULL$, which corresponds exactly to the statement of \cref{prop:dpinull_half_result}.
Hence, \cref{thm:step2stronger}  implies that $\DPI_0$ achieves at least half of the optimal improvement over $\NULL$.

We now prove \cref{thm:step2stronger}.
This proof makes use of the quantity $Z_{it}$, which was defined in the proof of \cref{prop:alg_minus_base}.
Fix any policy $\ALG$.
Recall that $I^{\ALG}_t$ and $I^*_t$ are the set of patients that are in state 0 under $\ALG$ and $\OPT$ respectively at time $t$.
Additionally, $D_t(I_t^{\ALG}) \subseteq I^{\ALG}_t$ are the patients that $\DPI_0$ would choose, out of patients in $I^{\ALG}_t$.
We first decompose the rewards based on whether or not a patient that was chosen in $\OPT$ was available to be chosen under $\ALG$.
\begin{align}
\OPT - \NULL
&= \bE\left[ \sumT \sum_{i \in A^*_t} Z_{it} \right]\nonumber \\
&= \bE\left[ \sumT \sum_{i \in A^*_t \cap I^{\ALG}_t} Z_{it} \right]
+ \bE\left[ \sumT \sum_{i \in A^*_t \setminus I^{\ALG}_t} Z_{it} \right].
\label{eq:opt_minus_base_decomp}
\end{align}

For the first term in \eqref{eq:opt_minus_base_decomp},
by definition of $D_t(I_t^{\ALG})$, we have
$\sum_{i \in A^*_t \cap I^{\ALG}_t} \nz_{it}(0) \leq \sum_{i \in D_t(I_t^{\ALG})} \nz_{it}(0)$.
Therefore, 
\begin{align*}
\bE\left[ \sumT \sum_{i \in A^*_t \cap I^{\ALG}_t} Z_{it} \right]
 = \bE\left[ \sumT \sum_{i \in A^*_t \cap I^{\ALG}_t} \nz_{it}(0) \right]
 \leq 
 \bE\left[ \sumT \sum_{i \in D_t(I_t^{\ALG})}  \nz_{it}(0) \right],
\end{align*}
where the first equality follows the same reasoning as \eqref{eq:propRandom}-\eqref{eq:propDet}.

\cref{thm:step2stronger} then follows from the following result which bounds the second term in \eqref{eq:opt_minus_base_decomp}. This term represents rewards from patients who are in state 0 under $\OPT$ but not under $\ALG$.
\begin{proposition} \label{prop:second_term}
\begin{align} \label{eq:second_term_claim}
\bE\left[ \sumT \sum_{i \in A^*_t \setminus I^{\ALG}_t} Z_{it} \right]	
\leq 
\bE\left[ \sumT \sum_{i \in A^{\ALG}_t} Z_{it} \right]	
\end{align}
\end{proposition}

\subsubsection{Proof of \cref{prop:second_term}.}
The main idea of this result is that for every $Z_{it}$ term that contributes to the LHS of \eqref{eq:second_term_claim}, since that patient is not available under $\ALG$, we have already collected this reward under $\ALG$.
We will show a one-to-one mapping from every $Z_{it}$ term on the LHS to a $Z_{i \nu_i(t)}$ term on the RHS of \eqref{eq:second_term_claim}.

Let $\cI = \{(i, t) : i \in A^*_t \setminus I^{\ALG}_t, K_{it} = 1\}$ be the set of (patient, time) tuples in which patient $i$ is chosen under $\OPT$ but not available under $\ALG$, and moreover, $K_{it} = 1$.
Note that from \eqref{eq:Z_explicit}, $K_{it} = 1$ is a necessary condition for $Z_{it} > 0$. 
Therefore, the LHS of \eqref{eq:second_term_claim} can be written as
\begin{align*}
\bE\left[ \sumT \sum_{i \in A^*_t \setminus I^{\ALG}_t} Z_{it} \right]	
= \bE\left[ \sum_{(i, t) \in \cI} Z_{it} \right].
\end{align*}

Fix $(i, t) \in \cI$.
We have that $S_{it}^{\ALG} = 1$ (since $i \notin I^{\ALG}_t$) and $S_{it}^{\OPT} = 0$ (since $i \in A^*_{t}$).
Define $\nu_i(t) < t$ to be the last time that $i$ was in state 0 under $\ALG$:
\begin{align} \label{eq:definetau}
\nu_i(t) = \max\{ t' < t : S_{it'}^{\ALG} = 0\}.
\end{align}
That is, under $\ALG$, patient $i$ transitioned from state 0 to state 1 between time $\nu_i(t)$ and $\nu_i(t)+1$ and stayed in state 1 since.
We can show that $\nu_i(t)$ satisfies the following property: it must be that $i$ was chosen at time $\nu_i(t)$ under $\ALG$, and moreover, $Z_{i \nu_i(t)}$ is at least $t - \nu_i(t)$.

\begin{claim} \label{claim:properties_of_y}
Let $(i, t) \in \cI$.
Then, $i \in A^{\ALG}_{\nu_i(t)}$, and $Z_{i \nu_i(t)} \geq t - \nu_i(t)$.
\end{claim}	
\begin{myproof}[Proof of \cref{claim:properties_of_y}]
Let $(i, t) \in \cI$.
By definition of $\nu_i(t)$, $S_{i, \nu_i(t)}^{\ALG} = 0$ and $S_{i, \nu_i(t)+1}^{\ALG} = 1$.
To the contrary, suppose $i \notin A_{\nu_i(t)}^{\ALG}$.
Then, $i$ transitioned to state 1 without being chosen, and therefore $P_{i\nu_i(t)} = 1$.
Then, it must be that $i$ transitions to state 1 under $\OPT$ as well; $S_{i, \nu_i(t)+1}^{\OPT} = 1$.
But since $S_{it}^{\OPT} = 0$, $i$ switches back to state 0 before state $t$, which means that this should also happen under $\ALG$ (due to the sample path coupling).
This is a contradiction by the definition of $\nu_i(t)$.
Therefore, $i \in A_{\nu_i(t)}^{\ALG}$.

Moreover, under the same reasoning, it must be that $K_{i\nu_i(t)} = 1$, and that $S_{is}^{\NULL} = 0$ for all $s \in \{\nu_i(t)+1, \dots, t\}$.
Therefore, $Z_{i \nu_i(t)} \geq t - \nu_i(t)$.
\end{myproof}

We next show that the mapping $\nu_i$ is one-to-one.
\begin{claim} \label{claim:1-1}
If $(i, t), (i, s) \in \cI$ for $t \neq s$, then $\nu_i(t) \neq \nu_i(s)$.
\end{claim}
\begin{myproof}[Proof of \cref{claim:1-1}]
Suppose $t < s$ such that $(i, t), (i, s) \in \cI$.
That is, under $\OPT$, patient $i$ was chosen both at time $t$ and $s$, and the patient was already in state 1 under $\ALG$ at both of these times.
Suppose, for contradiction, $\nu_i(t) = \nu_i(s)$.
This means that under $\ALG$, patient $i$ was chosen at time $\nu_i(t)$, and the patient has stayed in state 1 from time $\nu_i(t)+1$ until at least time $s$.

$(i, t), (i, s) \in \cI$ implies $S^{\OPT}_{it} = S^{\OPT}_{is} = 0$.
Since $K_{it} = 1$, the patient transitioned to state 1 at time $t+1$.
Since $S^{\OPT}_{is} = 0$, it must be that there exists a $t' \in \{t+1, \dots, s-1\}$ such that $G_{it'} = 1, P_{it'} = 0$.
But since $S^{\ALG}_{it'} = 1$, the fact that $G_{it'} = 1$ and $P_{it'} = 0$ implies that $S^{\ALG}_{i, t'+1} = 0$, which is a contradiction.
\end{myproof}

Claims \ref{claim:properties_of_y} and \ref{claim:1-1} show that every $(i, t) \in \cI$ maps to one term in the RHS of \eqref{eq:second_term_claim}.
Therefore, the right hand side of \eqref{eq:second_term_claim} is at least
\begin{align*}
\bE\left[ \sumT \sum_{i \in A^{\ALG}_t} Z_{it} \right]	
\geq 
\bE\left[ \sum_{(i, t) \in \cI} Z_{i\nu_i(t)} \right].
\end{align*}
Then, we are done if we can show
\begin{align*} 
\bE\left[ \sum_{(i, t) \in \cI} Z_{it} \right]
\leq
\bE\left[ \sum_{(i, t) \in \cI} Z_{i\nu_i(t)} \right],
\end{align*}
which we can write as
\begin{align}\label{eq:simpler_goal}
\sumT \sumN \bE[Z_{it} \;|\; (i, t) \in \cI\;] \Pr((i, t)\in \cI)
\leq 
\sumT \sumN \bE[Z_{i\nu_i(t)} \;|\; (i, t) \in \cI\;] \Pr((i, t)\in \cI).
\end{align}

The following claim implies \cref{eq:simpler_goal} and finishes the proof of \cref{prop:second_term}.
\begin{claim} \label{claim:compare_ys}
For every $i \in [N]$ and $t \geq 1$,
\begin{align} \label{eq:inner_term}
\bE[Z_{it} \;|\; (i, t)  \in \cI\;] 
\leq 
\bE[Z_{i\nu_i(t)} \;|\; (i, t)  \in \cI\;]
\end{align}
\end{claim}

\begin{myproof}[Proof of \cref{claim:compare_ys}]
Fix $i, t$.
First, we upper bound the LHS of \cref{eq:inner_term}.
\begin{align*}
\bE[Z_{it} \;|\; (i, t)  \in \cI\;]  
&= \bE[Z_{it} \;|\; Z_{it} \geq 1,  (i, t)  \in \cI\;] \Pr(Z_{it} \geq 1 \;|\; (i, t)  \in \cI) \\
&\leq \bE[Z_{it} \;|\; Z_{it} \geq 1,  (i, t)  \in \cI\;].
\end{align*}
Note that conditioned on $Z_{it} \geq 1$, $Z_{it}$ is only a function of $(P_{it'}, G_{it'})_{t' > t}$, the `future' with respect to $t$, and the event $\{(i, t)  \in \cI\} = \{i \in A^*_t \setminus I^{\ALG}_t,  K_{it} = 1\}$ is independent of these future random variables.
Therefore, conditioned on $Z_{it} \geq 1$, $Z_{it}$ is independent of $(i, t)  \in \cI$, and hence
\begin{align}\label{eq:lhs_ub} 
\bE[Z_{it} \;|\; (i, t)  \in \cI\;]  
\leq
\bE[Z_{it} \;|\; Z_{it} \geq 1].
\end{align}

Next, consider the RHS of \eqref{eq:inner_term}.
From \cref{claim:properties_of_y}, $(i, t) \in \cI$ implies $Z_{i \nu_i(t)} \geq t - \nu_i(t)$.
Therefore, 
\begin{align*}
\bE[Z_{i\nu_i(t)} \;|\; (i, t)  \in \cI\;]
= \bE[Z_{i\nu_i(t)} \;|\; (i, t)  \in \cI, Z_{i \nu_i(t)} \geq t - \nu_i(t)\;].
\end{align*}
Similar to before, conditioned on $Z_{i \nu_i(t)} \geq t - \nu_i(t)$, $Z_{it}$ is only a function of $(P_{it'}, G_{it'})_{t' > t}$, the `future' with respect to $t$.
Therefore, conditioned on $Z_{i \nu_i(t)} \geq t - \nu_i(t)$, $Z_{it}$ is independent of $(i, t) \in \cI$, and hence
\begin{align*}
\bE[Z_{i\nu_i(t)} \;|\; (i, t)  \in \cI\;]
&= \bE[Z_{i\nu_i(t)} \;|\; Z_{i \nu_i(t)} \geq t - \nu_i(t)\;] \\
&= \sum_{t' < t} \Pr(\nu_i(t) = t' \;|\; Z_{i \nu_i(t)} \geq t - \nu_i(t)) \bE[Z_{it'} \;|\; Z_{i t'} \geq t - t'] \\ 
&\geq \sum_{t' < t} \Pr(\nu_i(t) = t' \;|\; Z_{i \nu_i(t)} \geq t - \nu_i(t)) \bE[Z_{it'} \;|\; Z_{i t'} \geq 1].
\end{align*}

Note that for $s < t$, $\bE[Z_{is} | Z_{is} \geq 1] \geq \bE[Z_{it} | Z_{it} \geq 1]$.
This is because given the equation for $Z_{it}$ in \eqref{eq:Z_explicit}, the only difference between $Z_{it}$ and $Z_{is}$ is that $Z_{it}$ has a smaller maximum value of $T-t$.
Therefore,
\begin{align}
\bE[Z_{i\nu_i(t)} \;|\; (i, t)  \in \cI\;] 
&\geq  \bE[ Z_{i t}  | Z_{i t} \geq 1] \sum_{t' < t} \Pr(\nu_i(t) = t' \;|\; Z_{i \nu_i(t)} \geq t - \nu_i(t)) \nonumber \\
&= \bE[ Z_{it}  | Z_{it} \geq 1]. \label{eq:rhs_lb} 
\end{align}

Combining \eqref{eq:rhs_lb} and \eqref{eq:lhs_ub} proves the result.
\end{myproof}

\subsection{Step 3: Proof of \cref{prop:approx_index}}

Fix $\alpha_1, \alpha_2$,
and let $\ALG$ be a policy that uses indices that satisfies $\alpha_1 \nz_{it}(0) \leq z^{\ALG}_{it}(0) \leq \alpha_2 \nz_{it}(0)$.
Then, if patient $i$ and $j$ satisfy $z^{\ALG}_{it}(0) \geq z^{\ALG}_{jt}(0)$, then we have $\nz_{it}(0) \geq \frac{\alpha_1}{\alpha_2} \nz_{jt}(0)$.
Therefore, if one chooses patients with the highest values of $z^{\ALG}_{it}(0)$, their null intervention values will be at least an $\alpha_1/\alpha_2$ factor of the patients that have the highest null intervention values.
That is, at any time $t$, we have
\begin{align*}
\sum_{i \in A_t^{\ALG}} \nz_{it}(0) 
\geq 
\frac{\alpha_1}{\alpha_2} \sum_{i \in D_t(I_t^{\ALG}(0))} \nz_{it}(0),
\end{align*}
where $A_t^{\ALG}$ are the patients that $\ALG$ chooses at time $t$, $I_t^{\ALG}$ are the patients in state 0 at time $t$ under $\ALG$, and $D_t(I_t^{\ALG}) \subseteq I_t^{\ALG}$ are the $B$ patients with the largest null intervention values.
Summing over time steps and taking an expectation results in
\begin{align} \label{eq:comparez}
\bE\left[ \sumT \sum_{i \in A_t^{\ALG}} \nz_{it}(0) \right]
\geq 
\alpha \cdot \bE\left[ \sumT \sum_{i \in D_t(I_t^{\ALG})} \nz_{it}(0) \right].
\end{align}
\cref{prop:alg_minus_base} states that the LHS of \cref{eq:comparez} is equal to $\ALG - \NULL$.
Next, \cref{thm:step2stronger} implies that the RHS is at least $\frac{1}{2} (\OPT - \NULL)$.
Combining yields
\begin{align*}
\ALG - \NULL \geq \frac{\alpha}{2} (\OPT - \NULL)
\end{align*}
as desired.

\subsection{Step 4: Proof of \cref{lemma:rand_null}}  \label{s.app.p4}
Fix a patient $i$ and time $t$. In this proof, we remove the subscript $i$ on $p$, $g$ and $\tau$ for convenience.
\cref{lemma:zgamma} gives us an expression for both $z^{\lambda}_{it}(0)$ and $\nz_{it}(0)$:
\begin{align} 
z^{\lambda}_{it}(0) &= \tau \cdot \bE[\min\{\text{Geometric}(p + g + \gamma \tau (1-p-g)/(1-p)), T-t+1\}] \label{eq:z1}\\
\nz_{it}(0) &= \tau \cdot \bE[\min\{\text{Geometric}(p + g, T-t+1\}]. \label{eq:z2}
\end{align}
Since $p+g < 1$, $p + g + \gamma \tau (1-p-g)/(1-p)) > p+g$, hence $z^{\pi}_{it}(0)/\nz_{it}(0) \leq 1$.
Next, to lower bound the ratio $z^{\pi}_{it}(0)/\nz_{it}(0)$, we use the following lemma:
\begin{lemma} \label{lemma:trunc_geometric}
Suppose $X = \text{Geometric}(P)$, $Y = \text{Geometric}(G)$ with $P > G$, and let $T > 0$ be a positive integer.
Then, 
\begin{align*}
\frac{\bE[\min\{X, T\}]}{\bE[\min\{Y, T\}]} \geq \frac{\bE[X]}{\bE[Y]}.
\end{align*}
\end{lemma}

This lemma allows us to lower bound $z^{\pi}_{it}(0)/\nz_{it}(0)$ by considering the expressions \eqref{eq:z1} and \eqref{eq:z2} without the $\min$ with $T-t+1$.
That is, we have
\begin{align*}
\frac{\bE[\text{Geometric}(p + g + \gamma \tau(1-p-g)/(1-p) )]}{\bE[\text{Geometric}(p + g)]}
&= \frac{1}{1 + \gamma \frac{\tau (1-p-g)}{(p+g)(1-p)}}.
\end{align*}
Using \cref{lemma:trunc_geometric} implies
\begin{align*}
z^{\pi}_{it}/\nz_{it} 
&\geq \frac{1}{1 + \gamma \frac{\tau (1-p-g)}{(p+g)(1-p)}}.
\end{align*}

\begin{myproof}[Proof of \cref{lemma:trunc_geometric}]
Let $X = \text{Geometric}(P)$, $Y = \text{Geometric}(G)$ with $P > G$.
We can write an explicit expression for $\bE[\min\{X, T\}]$ as the following:
\begin{align*}
\bE[\min\{X, T\}] 
&= \sum_{k=1}^{T-1} k (1-P)^{k-1} P + T (1-P)^{T-1} \\
&= -P \frac{d}{dP} \left(\sum_{k=1}^{T-1} (1-P)^{k} \right) + T (1-P)^{T-1} \\
&= -P \frac{d}{dP} \left(\frac{1-(1-P)^T}{P}-1 \right) + T (1-P)^{T-1} \\
&= -P  \left(\frac{ T(1-P)^{T-1} P - (1-(1-P)^T) }{P^2}\right) + T (1-P)^{T-1} \\
&= \frac{1}{P} -  \frac{  (1-P)^T }{P} 
\end{align*}

Using this, we get the desired result:
\begin{align*}
	\frac{\bE[\min\{X, T\}] }{\bE[\min\{Y, T\}] } 
	&= \frac{\frac{1}{P} -  \frac{  (1-P)^T }{P}}{\frac{1}{G} -  \frac{  (1-G)^T }{G}} \\
	&= \frac{\frac{1}{P}(1-(1-P)^T)}{ \frac{1}{G}(1-(1-G)^T)} \\
	&\geq \frac{1/P}{1/G} \\
	&=\frac{\bE[X]}{ \bE[Y] },
\end{align*}
where the inequality follows since $P > G$.
\end{myproof}

\subsection{Proof of \cref{corr:approx}} \label{sec:app:pf_approx}

This result follows from applying \cref{prop:approx_index} and \cref{lemma:rand_null}.
Let $\ALG$ be a policy that satisfies \eqref{eq.approxvalues2}.
Then, by \cref{lemma:rand_null}, we have that the following holds for all $i$ and $t$:
\begin{align}
\frac{1}{1 + \gamma M_i} c_1 \nz_{it}(0) \leq z^{\ALG}_{it}(0)\leq c_2 \nz_{it}(0).
\end{align}
Then, we apply \cref{prop:approx_index} with $\alpha_1 = \frac{1}{1 + \gamma M_i} c_1$ and $\alpha_2 = c_2$, and the result follows.

\section{Proofs for \cref{ss.general_guarantees}} \label{s.app.decompbias}

\subsection{Proof of \cref{prop:dpi_can_be_worse}} \label{s.proof_dpi_worse}
We describe an instance where $\pi_0$ is an optimal policy, but $\DPI(\pi_0)$ is a suboptimal policy.
Consider the following instance of the two-state MDP model from \cref{ss.2statemodel}.
There are $N=3$ patients, where the first two patients have parameters $p_i = 0, \tau_i = 0.01, g_i = 0$ for $i=1, 2$, and the third patient has parameters $p_3 = 0, \tau_3 = 0.01, g_3 = 0.1$.
When patients 1 or 2 goes to state 1, they stay there indefinitely, while patient 3 does not (since $g_3 > 0$). 
Since all other parameters are the same, the value of an intervention is strictly higher for patient 1 and 2, versus patient 3.

Suppose all patients start at state 0, and there are $T=5$ time steps.
Let $\pi_0$ be a policy that assigns at most one intervention per time step defined using the following rules:
\begin{itemize}
  \item If both patients 1 and 2 are in state 0, assign it to one of them uniformly at random.
  \item Otherwise, if either patient 1 or 2 is in state 0, assign them an intervention.
  \item If neither patient 1 or 2 are in state 0, then assign an intervention to patient 3.
\end{itemize}
Notice that $\pi_0$ is the optimal policy.
However, $\DPI(\pi_0)$ ends up being a suboptimal policy.
At time $t=3$, the intervention value for patient 3 is $z_{33}^{\pi_0}(0)=0.029$, whereas the intervention value for patient 1 and patient 2 is $z_{13}^{\pi_0}(0)=z_{23}^{\pi_0}(0)=0.010$.
Therefore, at time $t=3$, the intervention value for patient 3 is higher than that of patient 1 or 2, and hence $\DPI(\pi_0)$ will (suboptimally) prioritize patient 3 at time 3.

The reason for this behavior is that under $\pi_0$, the only time that patient 3 receives an intervention at time 3 is in the unlikely event where both patient 1 and 2 are in state 1 at time 3.
When this event occurs, since patients 1 and 2 stay in state 1 indefinitely, patient 3 is also guaranteed to receive an intervention at time $t=4$ (if they did not switch to state 1 by then).
Then, the $q$-value for patient 3 at time 3, $q^{\pi_0}_{33}(s=0, a=1)$, 
incorporates the fact that they will \textit{also receive an intervention at time 4}.
Therefore, the intervention value, $z_{33}^{\pi_0}(0)$ effectively represents the increase in reward when patient 3 is given interventions at both time 3 and 4, and hence is higher than the intervention value for patient 1 and 2.
This behavior stems from the fact that $q^{\pi_0}_{33}(s=0, a=1)$ does not contain information regarding the fact that under $\pi_0$, patient 3 only receives an intervention under a very specific \textit{system} state.

\added{
\subsection{Proof of \cref{thm:improvement_random}} \label{s.proof_improvement_random}
Let $\pi_0 = \RAND(\gamma)$ and let $\pi_1 = \DPI(\RAND(\gamma))$.
For any system policy $\pi$, let $V_t^{\pi}$ and $Q_t^{\pi}$ denote the system value function and the system Q-function respectively for the system MDP under policy $\pi$:
\begin{align*} 
V_{t}^\pi(\tilde{\bS}) &= \bE_{\pi}\left[\sum_{t'=t}^T \sumN R(S_{it'}^\pi, S_{i,t'+1}^\pi, A_{it'}^\pi) \;\bigg|\; \bS_t^\pi = \tilde{\bS} \right], \\
Q_{t}^\pi(\tilde{\bS}, \tilde{\bA}) &= \bE_{\pi}\left[\sum_{t'=t}^T \sumN R(S_{it'}^\pi, S_{i,t'+1}^\pi, A_{it'}^\pi) \;\bigg|\;  \bS_t^\pi = \tilde{\bS},  \bA_t^\pi = \tilde{\bA} \right].
\end{align*}
With this notation, we have that $\RAND(\gamma) = V_1^{\pi_0}(\bS_1)$ and $\DPI(\RAND(\gamma)) = V_1^{\pi_1}(\bS_1)$, where $\bS_1$ is the initial state that specified by the instance.
Our goal is to show $V_1^{\pi_1}(\bS_1) \geq V_1^{\pi_0}(\bS_1)$.

Note that under the policy $\pi_0 = \RAND(\gamma)$, the actions $A_{it}$ are independently chosen for each patient $i$. 
Therefore, the system value function and the Q-functions can be decomposed as:
\begin{align*} 
V_{t}^{\pi_0}(\tilde{\bS}) = \sum_{i=1}^N \big(\gamma q_{it}^{\pi_0}(\tilde{S}_{i}, 1) + (1-\gamma) q_{it}^{\pi_0}(\tilde{S}_{i}, 0) \big), \quad
Q_{t}^{\pi_0}(\tilde{\bS}, \tilde{\bA}) = \sum_{i=1}^N q_{it}^{\pi_0}(\tilde{S}_{i}, \tilde{A}_{i}).
\end{align*}
For a system state $\bS$, let $\bA_t^{\pi_1}(\bS) \in \{0, 1\}^N$ be the action that $\pi_1$ chooses at state $\bS$ at time $t$.
We claim that under any state, it is better to take the action from $\pi_1$ and then continue with $\pi_0$ henceforth, compared to just following $\pi_0$.
\begin{claim} \label{claim:improvement}
For any state $\tilde{\bS}$ and any $t$, $Q_{t}^{\pi_0}(\tilde{\bS}, \bA_t^{\pi_1}(\tilde{\bS}) ) \geq V_{t}^{\pi_0}(\tilde{\bS})$.
\end{claim}

\begin{myproof}[Proof of \cref{claim:improvement}]
Recall that $z^{\pi_0}_{it}(s) = q_{it}^{\pi_0}(s, 1) - q_{it}^{\pi_0}(s, 0)$.
Then, we can write the value function $V_{t}^{\pi_0}(\tilde{\bS})$ as
\begin{align*}
V_{t}^{\pi_0}(\tilde{\bS}) &= \sum_{i=1}^N q_{it}^{\pi_0}(\tilde{S}_{i}, 0) + \gamma \sum_{i=1}^N  z^{\pi_0}_{it}(\tilde{S}_i).
\end{align*}
Similarly, we can write $Q_{t}^{\pi_0}(\tilde{\bS}, \bA_t^{\pi_1}(\tilde{\bS}))$ as
\begin{align*}
Q_{t}^{\pi_0}(\tilde{\bS}, \bA_t^{\pi_1}(\tilde{\bS})) &= \sum_{i=1}^N q_{it}^{\pi_0}(\tilde{S}_{i}, 0) + \sum_{i: \bA_t^{\pi_1}(\tilde{\bS})_i =1}  z^{\pi_0}_{it}(\tilde{S}_i).
\end{align*}
Note that by definition of $\pi_1$, $\bA_t^{\pi_1}(\tilde{\bS})_i =1$ for patients $i$ with the $B$ highest values of $z^{\pi_0}_{it}(\tilde{S}_i)$.
Since $B \geq \gamma N$, we have that $\sum_{i: \bA_t^{\pi_1}(\tilde{\bS})_i =1}  z^{\pi_0}_{it}(\tilde{S}_i) \geq \gamma \sum_{i=1}^N  z^{\pi_0}_{it}(\tilde{S}_i)$.
Therefore, $Q_{t}^{\pi_0}(\tilde{\bS}, \bA_t^{\pi_1}(\tilde{\bS})) \geq V_{t}^{\pi_0}(\tilde{\bS})$.
\end{myproof}

We leverage the claim to show that $V_1^{\pi_0}(\bS) \leq V_1^{\pi_1}(\bS)$ for any $\bS$, which finishes the proof.
\begin{align*}
V_1^{\pi_0}(\bS) 
&\leq Q_{1}^{\pi_0}(\bS, \bA^{\pi_1}(\bS)) \\
&=\bE_{\bS' \sim P(\bS, \cdot, \pi_1)}\left[ \sum_{i=1}^N R_i(S_{i}, S'_{i}, \bA_1^{\pi_1}(\bS)_i) +  V_{2}^{\pi_0}(\bS') \right] \\
&\leq \bE_{\bS' \sim P(\bS, \cdot, \pi_1)}\left[ \sum_{i=1}^N R_i(S_{i}, S'_{i}, \bA_1^{\pi_1}(\bS)_i) +  Q_{2}^{\pi_0}(\bS', \bA^{\pi_1}(\bS')) \right] \\
&= \bE_{\bS' \sim P(\bS, \cdot, \pi_1), \bS'' \sim P(\bS', \cdot, \pi_1)}\left[ \sum_{i=1}^N R_i(S_{i}, S'_{i}, \bA_1^{\pi_1}(\bS)_i) +  \sum_{i=1}^N R_i(S'_{i}, S''_{i}, \bA_2^{\pi_1}(\bS')_i) +  V_{3}^{\pi_0}(\bS'') \right] \\
&\leq \ldots \\
&\leq \bE_{\bS' \sim P(\bS, \cdot, \pi_1), \bS'' \sim P(\bS', \cdot, \pi_1), \dots}\left[ \sum_{i=1}^N R_i(S_{i}, S'_{i}, \bA_1^{\pi_1}(\bS)_i) +  \sum_{i=1}^N R_i(S'_{i}, S''_{i}, \bA_2^{\pi_1}(\bS')_i) +  \dots \right] \\
&= V^{\pi_1}_1(\bS).
\end{align*}
}

\section{Synthetic Simulations} \label{s.app.synthetic simulations}

We run a set of synthetic simulations for the two-state model of \cref{s.theory}.
We use 1000 patients and 500 time steps.
For each patient $i$, we generate their parameters independently at random as 
$p_i, g_i, \tau_i \sim \text{Unif}(0, 0.2)$.
We vary the budget as $B \in \{5, 10, 50, 100, 300\}$ and we try four policies:
\begin{itemize}
\item $\NULL$: Does not give any interventions.
\item $\baseline$: Gives interventions to $B$ patients uniformly at random, out of those in state 0.
\item $\DPI(0)$: Our algorithm which prioritizes patients based on the value of $\tau_i/(p_i + g_i)$ (see \cref{prop:z_clean_form}). 
\item $\textsf{Whittle}$: Prioritizes patients based on the value of $\frac{p_i+\tau_i}{g_i} -  \frac{p_i(p_i + g_i + \tau_i)}{g_i(p_i + g_i)}$, which corresponds to the Whittle's index for patient $i$ under the two-state model.
\end{itemize}
The results, given as a percentage improvement over the $\NULL$ policy, can be found in \cref{tab.synthetic_results}.

\begin{table}[h]
\TableSpaced %
\caption{
The average improvement in total reward over the $\NULL$ policy which does not assign any interventions.
There were 1000 patients in total, whose parameters were generated at random.
} \label{tab.synthetic_results}
\vspace{2mm}
\begin{center}
\begin{tabular}{@{}c|cccc@{}}
\toprule
Budget & \quad $\baseline$ (random) \;  & $\DPI(0)$ & $\textsf{Whittle}$  \\ \midrule
5      &  0.72\%  & 3.35\%   & 3.30\%  \\
10     &  1.45\%  & 5.10\%   & 5.15\%  \\ 
50     &  6.66\%  & 13.99\%  & 13.95\%   \\
100    &  12.40\% & 21.13\%  & 21.10\%   \\
300    &  31.76\% & 34.46\%  & 34.46\%   \\ \bottomrule
\end{tabular}
\end{center}
\end{table}

We see that the performance of $\DPI$ and $\textsf{Whittle}$ are almost identical, while they both significantly outperform $\baseline$.
We also observe that the relative improvement over $\baseline$ decreases in the budget, which means that carefully targeting patients is more important when the budget is small. When the budget is large, many patients can get the intervention, so  careful targeting is not as critical.

\section{Details on Keheala Case Study} \label{s.app.keheala}

\subsection{List of features} \label{s.app.list_features}

\textbf{Static features.}
For each patient, we include the following static covariates: weight, height, age, sex, language, county, HIV positive, and extrapulmonary TB.
There were 6 different counties where we used a one-hot encoding, which resulted in 13 features in total.

\textbf{Condensed history.}
For patient $i$ at time $t$, we summarize their history, $H_{it} = (V_{i1}, A_{i1}, \dots, V_{i, t-1}, A_{i,t-1}, V_{it}) \in \bR^{2t-1}$, using the following features:
\begin{itemize}
	\item Verifications: total so far, total percentage, total last week, X days ago for the last $X \in \{1, \dots, 7\}$ days.
	\item Verification/non-verification streaks (how many days in a row a patient verifies / does not verify): current streak, longest streak
	\item Interventions: total so far, total last week, X days ago for the last $X \in \{1, 2, 3\}$ days.
	\item Number of days on the platform, number of days of treatment left.
\end{itemize}
This results in 21 features in total for the condensed history.
A similar structure of features was used in \cite{Boutilier22Improving} which analyzed the same dataset.

\subsection{Simulation Model Details} \label{s.app.simulation}

We briefly describe the double ML method of \cite{chernozhukov2018double}.
Let $Y \in \bR$ be the outcome variable, $T \in \{0, 1\}$ the treatment, and $X \in \bR^{d}$ the observable features.
The model makes the following structural assumptions:
\begin{align*}
Y &= \tau(X) \cdot T + g(X) + \eps, \\ 
T &= f(X) + \eta,
\end{align*}
where $\bE[\eps | X] = 0, \bE[\eta | X] = 0$, and $\bE[\eps \cdot \eta | X] = 0$.
The goal is to estimate the conditional average treatment effect, $\tau(X)$.
We estimate two functions:
\begin{align*}
q(X) = \bE[Y \;|\; X], \quad
f(X) = \bE[T \;|\; X].
\end{align*}
Then, we compute the residuals
\begin{align*}
\tilde{Y} = Y - q(X), \quad \tilde{T} = T - f(X).
\end{align*}
Lastly, we estimate
\begin{align*}
\hat{\tau} = \argmin_{\tau} \bE[(\tilde{Y} - \tau(X) \cdot \tilde{T})^2].
\end{align*}

For the Keheala case study, we used gradient boosting to estimate $q$ and $f$, and we assumed a linear function for $\tau(X) = \theta^\top X$.

Lastly, we cap $\hat{\tau}(s)$ so that the resulting function $f(s, 1)$ is between 0 and 1.
That is, we let $\tau(s) = \max\{\min\{\hat{\tau}(s), 1 - f(s, 0)\}, -f(s, 0)\}$.

\subsection{Details on the Bandit Algorithm} \label{sec:app:bandit}
The pseudocode for the bandit algorithm used in the experiments can be found in \cref{alg:bandit}.
The bandit is initialized with the offline dataset, the same data that was used to train $\DPI$ -- the $\mathbf{S}(a)$  and $\mathbf{V}(a)$ that is an input in \cref{alg:bandit} comes from the offline data. 
The algorithm uses a Thompson Sampling approach, where a parameter is sampled from the posterior, and the intervention is given to those with the highest intervention values with respect to the sampled parameter. 
In our experiments, we used $\sigma^2 = 1/4$ \added{(we tried several different parameters values and chose the best one)}. 

\begin{algorithm}
\caption{Thompson Sampling Bandit Policy}
\label{alg:bandit}
\begin{algorithmic}[1] %
\Require Budget $B$, $\mathbf{S}(a), \mathbf{V}(a)$ for $a \in \{0, 1\}$, noise $\sigma^2$.
	\For{$t=1, \dots, T$}
	\For{$a \in \{0, 1\}$}
		\State $\hat{\beta}(a) = (\mathbf{S}(a)^{\top} \mathbf{S}(a))^{-1} \mathbf{S}(a)^{\top} \mathbf{V}(a)$ 
		\State $\hat{\Sigma}(a) = \sigma^2 (\mathbf{S}(a)^{\top} \mathbf{S}(a))^{-1}$
		\State $\tilde{\beta}(a) \sim \text{Normal}(\hat{\beta}(a), \hat{\Sigma}(a))$
			\Comment Sample parameter from posterior
	\EndFor
	\For{$i=1, \dots, N$}
	\State $\tilde{z}_{it} = (\tilde{\beta}(1)-\tilde{\beta}(0))^{\top} S_{it}$
	\Comment Sampled intervention values for each patient
	\EndFor
	\State Assign interventions to $B$ patients with the largest $\tilde{z}_{it}$ values.
	\For{$i=1, \dots, N$}
		\Comment Update parameters
		\State $V_{i,t+1} = 1 \text{ if patient $i$ verified at $t+1$, otherwise } 0$
		\If{$A_{it} = 1$}
		\State $\mathbf{S}(1) = \text{Append}(\mathbf{S}(1), S_{it})$
		\State $\mathbf{V}(1) = \text{Append}(\mathbf{V}(1), V_{i,t+1})$
		\Else
		\State $\mathbf{S}(0) = \text{Append}(\mathbf{S}(0), S_{it})$
		\State $\mathbf{V}(0) = \text{Append}(\mathbf{V}(0), V_{i,t+1})$
		\EndIf
    \EndFor
    \EndFor
 \end{algorithmic}
\end{algorithm}

\added{We note that action-centered contextual bandits \citep{greenewald2017action} is a variant of Thompson Sampling that has been used in a similar setting (e.g., HeartSteps study of \cite{liao2020personalized}).
However, this method does not readily apply to our setting due to the budget constraint, hence we use \cref{alg:bandit}.
}

\subsection{Details on the QWI algorithm} \label{sec:app:qwi}

\paragraph{Learning the index.}
For a given $\lambda \in \bR$, we consider a process in which every time action 0 is taken, a reward of $\lambda$ is awarded.
We define $Q(S, A; \lambda)$ as the Q-value from state $S$ and action $A$ under this process.
Then, the $Q$-values satisfy:
\begin{align}
	Q(S, A; \lambda) = r(S, A) + (1-A) \lambda - \beta + \bE_{S' \sim P(S, \cdot, A)}[\max_{a'} Q(S', A'; \lambda)],
\end{align}
where $\beta \in \bR$ is the optimal reward.
Then, the Whittle's index for state $s$ is the value $\lambda(s)$ that satisfies
\begin{align}
	Q(S, 0; \lambda(S)) = Q(S, 1; \lambda(S)).
\end{align}

Recall that we have offline samples $\{(S_i, A_i, v_i, \ell_i, S_i')\}$, where $v_i \in \{0, 1\}$ is the outcome of whether the patient verified, $\ell_i \in \bN$ is the number of days the patient has left, and $S_i'$ is the next state of the patient.
Fix a number $\lambda$, and we will aim to learn $Q(S, A; \lambda)$, which represents the future verification rate.
Initialize $Q_0(\cdot, \cdot; \lambda) = 0$.
Let $\cI(S, A) = \{i : S_i = S, A_i = A\}$ be the data points with state $S$ and action $A$.
For $t = 1, 2, \dots$, update the $Q$-values as:
\begin{align} \label{eq:q_iteration}
Q_t(S, A; \lambda) = \frac{1}{|\cI(S, A)|} \sum_{i \in \cI(S, A)} \frac{1}{\ell_i} \big(v_i  + (1-A_i) \lambda + (\ell_i-1) \max_{a'}Q_{t-1}(S_i', A')\big).
\end{align}	
Then, for every state $u$, to find the $\lambda(S)$ such that $Q(S, 0; \lambda(S)) = Q(S, 1; \lambda(S))$, 
we did a binary search across $\lambda$. 
At each iteration of the binary search, for a given value of $\lambda$, we ran the above method \eqref{eq:q_iteration} for $t=100$ iterations.

\subsection{Details on Prominent Features for $\DPI$} \label{s.app.coefficients}
We describe how \cref{tab.coefs} was generated. 
First, we compute $\hat{\theta}_0$ using the least square regression in \eqref{eq:leastsquares}.
Then, for each sample $(S_{it}, A_{it}, y_{it})$ where $A_{it} = 1$, we create a new target $\tilde{y}_{it} = y_{it} - \hat{\theta}_0^{\top} S_{it}$, which simply subtracts off the prediction from the first regression.
Then, the intervention value is essentially the regression of $S_{it}$ onto the new target values.
Instead of performing the usual least-squares, we first normalize each feature so that they have a standard deviation of 1, and then we perform a Lasso regression \citep{tibshirani1996regression}:
\begin{align*} 
	\tilde{\theta} &\in \argmin_{\theta \in \bR^{34}} \bigg( \sum_{i \in N} \sum_{t=T_{\text{s}}(i)}^{T_{\text{e}}(i)}  \bI(A_{it} = 1)(\tilde{y}_{it} - \theta^\top \tilde{S}_{it})^2 + \lambda ||\theta||^2_1 \bigg),
\end{align*}
where $\tilde{S}$ represent the states after column normalization.
We chose the value of $\lambda$ so that the output $\tilde{\theta}$ has five non-zero entries.
These are the features and their respective sign of the coefficient that is shown in \cref{tab.coefs}.

\end{document}